\documentclass[10pt,twocolumn,letterpaper]{article}

\usepackage{cvpr}              

\usepackage{graphicx}
\usepackage{amsmath}
\usepackage{amssymb}
\usepackage{booktabs}

\usepackage[pagebackref,breaklinks,colorlinks]{hyperref}

\usepackage[capitalize]{cleveref}
\crefname{section}{Sec.}{Secs.}
\Crefname{section}{Section}{Sections}
\Crefname{table}{Table}{Tables}
\crefname{table}{Tab.}{Tabs.}

\def\cvprPaperID{4656} 
\def\confName{CVPR}
\def\confYear{2023}

\usepackage{color}
\usepackage{xcolor}
\def\red#1{\textcolor[rgb]{1,0,0}{#1}}

\newcommand{\keypoint}[1]{\vspace{0.1cm}\noindent\textbf{#1}\;}
\newcommand{\cut}[1]{}
\usepackage{multirow}
\usepackage{amsmath}
\usepackage{amssymb}
\usepackage{mathtools}
\usepackage{arydshln}
\usepackage{booktabs}
\usepackage{colortbl}
\usepackage{stmaryrd}
\definecolor{Gray}{gray}{0.9}
\makeatletter
\apptocmd\@maketitle{{\myfigure{}\par}}{}{}
\makeatother
\usepackage{capt-of,etoolbox}

\begin{document}

\title{\textit{Picture that Sketch}: Photorealistic Image Generation from Abstract Sketches}

\author{Subhadeep Koley\textsuperscript{1,2} \hspace{.2cm}  Ayan Kumar Bhunia\textsuperscript{1} \hspace{.2cm} Aneeshan Sain\textsuperscript{1,2} \hspace{.2cm}  Pinaki Nath Chowdhury\textsuperscript{1,2} \\
Tao Xiang\textsuperscript{1,2}\hspace{.3cm}  Yi-Zhe Song\textsuperscript{1,2} \\
\textsuperscript{1}SketchX, CVSSP, University of Surrey, United Kingdom.  \\
\textsuperscript{2}iFlyTek-Surrey Joint Research Centre on Artificial Intelligence.\\
{\tt\small \{s.koley, a.bhunia, a.sain, p.chowdhury, t.xiang, y.song\}@surrey.ac.uk}
}
\vspace{-0.4cm}
\newcommand\myfigure{
\centering
\vspace{-0.9cm}
\captionsetup{type=figure} 
    \includegraphics[width=\textwidth]{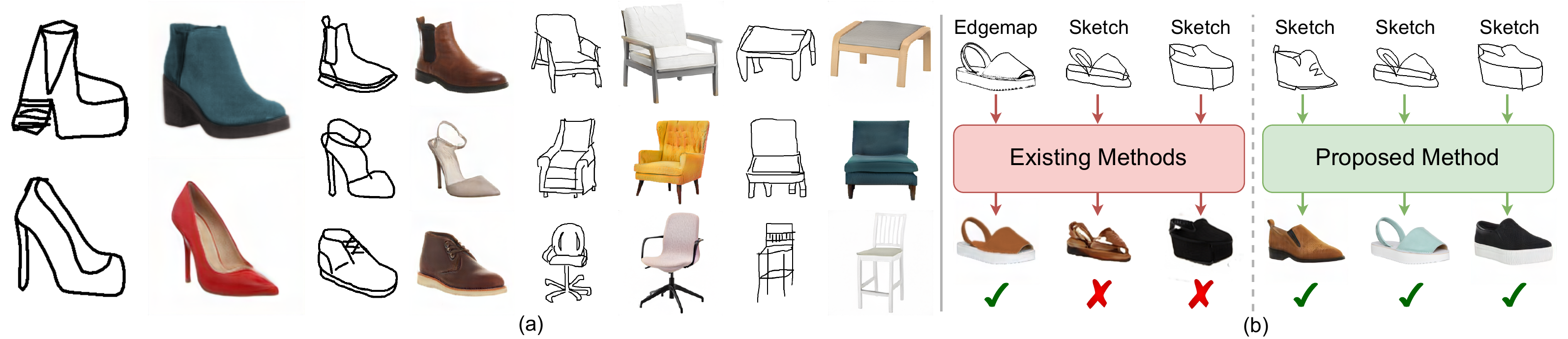}
    \vspace{-0.8cm}
\captionof{figure}{(a) Set of photos generated by the proposed method. (b) While existing methods can generate faithful photos from perfectly pixel-aligned \textit{edgemaps}, they fall short drastically in case of highly deformed and sparse \textit{free-hand sketches}. In contrast, our autoregressive sketch-to-photo generation model produces highly photorealistic outputs from highly \textit{abstract sketches}.}
\label{fig:teaser}
\vspace{+0.1cm}
}

\maketitle
\begin{abstract}
\vspace{-0.3cm}

Given an abstract, deformed, ordinary sketch from untrained amateurs like you and me, this paper turns it into a photorealistic image -- just like those shown in \cref{fig:teaser}(a), all non-cherry-picked. We differ significantly from prior art in that we do not dictate an edgemap-like sketch to start with, but aim to work with abstract free-hand human sketches. In doing so, we essentially democratise the sketch-to-photo pipeline, ``picturing'' a sketch regardless of how good you sketch. Our contribution at the outset is a decoupled encoder-decoder training paradigm, where the decoder is a StyleGAN trained on photos only. This importantly ensures that generated results are always photorealistic. The rest is then all centred around how best to deal with the abstraction gap between sketch and photo. For that, we propose an autoregressive sketch mapper trained on sketch-photo pairs that maps a sketch to the StyleGAN latent space. We further introduce specific designs to tackle the abstract nature of human sketches, including a fine-grained discriminative loss on the back of a trained sketch-photo retrieval model, and a partial-aware sketch augmentation strategy. Finally, we showcase a few downstream tasks our generation model enables, amongst them is showing how fine-grained sketch-based image retrieval, a well-studied problem in the sketch community, can be reduced to an image (generated) to image retrieval task, surpassing state-of-the-arts. We put forward generated results in the supplementary for everyone to scrutinise. Project page: \url{https://subhadeepkoley.github.io/PictureThatSketch}
\end{abstract}

\vspace{-0.4cm}
\section{Introduction}
\vspace{-0.1cm}
People sketch, some better than others. Given a shoe image like ones shown in \cref{fig:teaser}(a), everyone can scribble a few lines to depict the photo, again mileage may vary -- top left sketch arguably lesser than that at bottom left. The opposite, \textit{i.e.}, hallucinating a photo based on even a very abstract sketch, is however something humans are very good at having evolved on the task over millions of years. This seemingly easy task for humans, is exactly one that this paper attempts to tackle, and apparently does fairly well at -- given an abstract sketch from untrained amateurs like us, our paper turns it into a photorealistic image (see \cref{fig:teaser}).

This problem falls into the general image-to-image translation literature~\cite{isola2017image, pang2021image}. Indeed, some might recall prior arts (\textit{e.g.}, pix2pix~\cite{isola2017image}, CycleGAN~\cite{zhu2017unpaired}, MUNIT~\cite{huang2018multimodal}, BicycleGAN~\cite{zhu2017toward}), and sketch-specific variants~\cite{ham2022cogs, wang2021sketch} primarily based on pix2pix~\cite{isola2017image} claiming to have tackled the exact problem. We are strongly inspired by these works, but significantly differ on one key aspect -- we aim to generate from abstract human sketches, not accurate photo edgemaps which are already ``photorealistic''.

This is apparent in \cref{fig:teaser}(b), where when edgemaps are used prior works can hallucinate high-quality photorealistic photos, whereas rather ``peculiar'' looking results are obtained when faced with amateur human sketches. This is because all prior arts assume pixel-alignment during translation -- so your drawing skill (or lack of it), got accurately reflected in the generated result. As a result, chance is you and me will not fetch far on existing systems if not art-trained to sketch photorealistic edgemaps -- we, in essence, democratise the sketch-to-photo generation technology, ``picturing'' a sketch regardless of how good you sketch.

Our key innovation comes after a pilot study where we discovered that the pixel-aligned artefact~\cite{richardson2021encoding} in prior art is a direct result of the typical encoder-decoder~\cite{isola2017image} architecture being trained end-to-end -- this enforces the generated results to strictly follow boundaries defined in the input sketch (edgemap). Our first contribution is therefore a decoupled encoder-decoder training, where the decoder is pre-trained StyleGAN~\cite{karras2019style} trained on photos only, and is frozen once trained. This importantly ensures generated results are sampled from the StyleGAN~\cite{karras2019style} manifold therefore of photorealistic quality.

The second, perhaps more important innovation lies with how we bridge the abstraction gap~\cite{hertzmann2020line, chowdhury2023scenetrilogy, chowdhury2023what} between sketch and photo. {For that, we propose to train an encoder that performs a mapping from abstract sketch representation to the latent space of the learned latent space of StyleGAN~\cite{karras2019style} (\textit{i.e.}, not actual photos as per the norm).} To train this encoder, we use ground-truth sketch-photo pairs, and impose a novel fine-grained discriminative loss between the input sketch and the generated photo, together with a conventional reconstruction loss~\cite{zhang2018unreasonable} between the input sketch and the ground-truth photo, to ensure the accuracy of this mapping process. To double down on dealing with the abstract nature of sketches, we further propose a partial-aware augmentation strategy where we render partial versions of a full sketch and allocate latent vectors accordingly (the more partial the input, the lesser vectors assigned).

Our autoregressive generative model enjoys a few interesting properties once trained: \textit{(i)} abstraction level (\textit{i.e.}, how well the fine-grained features in a sketch are reflected in the generated photo) can be easily controlled by altering the number of latent vectors predicted and padding the rest with Gaussian noise, \textit{(ii)} robustness towards noisy and partial sketches, thanks to our partial-aware sketch augmentation strategy, and \textit{(iii)} good generalisation on input sketches across different abstraction levels (from edgemaps, to sketches across two datasets). We also briefly showcase two potential downstream tasks our generation model enables: fine-grained sketch-based image retrieval (FG-SBIR), and precise semantic editing. On the former, we show how FG-SBIR, a well-studied task in the sketch community~\cite{bhunia2022sketching, sain2022sketch3t, sain2023exploiting, sain2023clip}, can be reduced to an image (generated) to image retrieval task, and that a simple nearest-neighbour model based on VGG-16~\cite{simonyan2014very} features can already surpass state-of-the-art. On the latter, we demonstrate how precise local editing can be done that is more fine-grained than those possible with text and attributes.

We evaluate using conventional metrics (FID, LPIPS), plus a new retrieval-informed metric to demonstrate superior performance. But, as there is no better way to convince the jury other than presenting \textit{all} facts, we offer \textit{all} generated results in the supplementary for everyone to scrutinise.

\vspace{-0.2cm}
\section{Related Works}
\vspace{-0.25cm}
\noindent \textbf{Image-to-Image Translation:} Images from source domain can be translated to a specific target domain through a learned generative mapping function, to perform tasks like semantic label-map to RGB~\cite{park2019semantic}, day-to-night~\cite{isola2017image}, edgemap-to-photo~\cite{isola2017image} translations. Following the advent of deep neural networks, the seminal work of pix2pix ~\cite{isola2017image} introduced a unified framework that trains a U-Net-based generator~\cite{ronneberger2015u} with a weighted summation of reconstruction and adversarial GAN losses~\cite{isola2017image}.
It essentially generates a pixel-to-pixel mapped output $I'(x,y)$ in the target domain corresponding to input $I(x,y)$ from source domain~\cite{isola2017image}. This has consequently laid foundation to various vision tasks, like image colourisation~\cite{xiao2019single}, conditional image generation~\cite{ huang2018multimodal, choi2020stargan},   style-transfer~\cite{zhu2017unpaired}, inpainting~\cite{han2019finet} and enhancements~\cite{zhu2017unpaired, qu2019enhanced, ledig2017photo}. Furthermore, pix2pix first illustrated generation of pixel-perfect photos~\cite{isola2017image} even from sparse line drawings like edgemaps. However, making it work for \textit{free-hand sketch} is still an open problem as sketch is highly abstract~\cite{lu2018image}, lacking alignment, unlike edgemaps.

\vspace{0.05cm}
\noindent \textbf{Sketch-to-Photo Generation:} Photorealistic image (photo) generation from free-hand sketches is still in its infancy, despite significant advances on various sketch-based vision tasks~\cite{sain2022sketch3t, bhunia2022doodle, wang2021sketchembednet, xie2021exploiting, bhunia2023sketch2saliency}. Pix2pix~\cite{isola2017image} forms the basis for most of the recent deep learning-based sketch-to-photo generation frameworks (\Cref{tab:relatedWorks}). Particularly they use either GAN-based models~\cite{chen2018sketchygan, gao2020sketchycoco} with conditional self-attention~\cite{li2019linestofacephoto}, feature manifold projection~\cite{chen2020deepfacedrawing}, domain adaptation~\cite{xiang2022adversarial}, two-stage generation~\cite{ghosh2019interactive}, or contextual loss~\cite{lu2018image}. Nonetheless, the majority of these works~\cite{li2019linestofacephoto, chen2020deepfacedrawing} are restricted to using edgemaps as a pseudo sketch-replacement for model training. However, a \textit{free-hand} sketch~\cite{chowdhury2022fs} with human-drawn sparse and abstract strokes, is a way of conveying the ``semantic intent'', and largely differs~\cite{lu2018image} from an edgemap. While edgemap perfectly aligns with photo boundaries, a sketch is a human abstraction of any object/concept, usually with strong deformations~\cite{lu2018image}. To alleviate this, earlier attempts have been made via unsupervised training~\cite{liu2020unsupervised, yi2017dualgan} by excluding paired sketch-photo data, or using specific loss functions~\cite{lu2018image}. The generated images nevertheless follow the sketch boundaries, yielding deformed photos.

\begin{table*}[!htbp]
\renewcommand{\arraystretch}{0.8}
\scriptsize
\centering

\caption{Recent sketch-to-photo generation literature can be grouped as -- \textit{(i)} Categorical, \textit{(ii)} Semi Fine-Grained (FG), \textit{(iii)} Scene-level, and \textit{(iv)} Facial-Photo. Additionally, we summarise existing state-of-the-arts in terms of training data preparation and salient design choices.}
\vspace{-0.3cm}
\begin{tabular}{c c c l}
\toprule
\textbf{Paper} & \textbf{Category} & \textbf{Type of Sketch} & \multicolumn{1}{c}{\cellcolor{white!30}\textbf{Data Preparation + Salient Design Component}}\\\midrule
SketchyGAN~\cite{chen2018sketchygan} & Categorical & Synth+Real & \textbullet~Fully automatic edgemap augmentation \textbf{+} input injection at multiple layers. \\
iSketch\&Fill~\cite{ghosh2019interactive} & Categorical & Synthetic & \textbullet~Edgemap creation with Im2Pencil~\cite{li2019im2pencil} and sketch-simplification~\cite{simo2016learning} \textbf{+} ResNet~\cite{he2016deep} generator.\\
CoGS~\cite{ham2022cogs} & Categorical & Synth+Real & \textbullet~Saliency with~\cite{hwang2021exemplar} for synthetic sketches \textbf{+} VQ-GAN~\cite{esser2021taming} with VAE over codebook vectors. \\\midrule
ContextGAN~\cite{lu2018image} & Semi-FG & Synthetic & \textbullet~Synthetic sketch generation with XDoG~\cite{winnemoller2012xdog} and~\cite{kang2007coherent} \textbf{+} optimisation-based GAN inversion.\\
Two-Stage~\cite{liu2020unsupervised} &  Semi-FG  & Real & \textbullet~Synthetic noisy-stroke for augmentation \textbf{+} two-stage sketch-to-edgemap-to-photo generation. \\
SYO-GAN~\cite{wang2021sketch} & Semi-FG & Synthetic & \textbullet~Pseudo sketch creation with PhotoSketch~\cite{li2019photo} \textbf{+} Fine-tuning GAN model with a few pose-specific sketches.\\\midrule
SketchyCOCO~\cite{gao2020sketchycoco} & Scene-level & Semi-real & \textbullet~Synthetic scene sketch \textbf{+} generation of foreground object followed by contextual background. \\
Two-Stage~\cite{wang2022unsupervised} & Scene-level & Synthetic & \textbullet~Edgemaps generated with~\cite{poma2020dense} \textbf{+} edgemap standardisation followed by content-style disentanglement.\\ \midrule
DeepFaceDraw~\cite{chen2020deepfacedrawing} & Facial photo & Synthetic & \textbullet~Photocopy filter and~\cite{simo2016learning} for training data \textbf{+} region-wise embedding with 2-stage generation. \\
Controlled S2I~\cite{yang2021controllable} & Facial photo & Synthetic & \textbullet~HED~\cite{xie2015holistically} for edgemaps + dilation-based sketch refinement network for adapting edge-based models. \\
\rowcolor{Gray}
\textbf{Proposed} & \textbf{Fine-grained} & Real & \textbullet~Unlabelled photos \& sketch-photo pairs \textbf{+} autoregressive latent-mapper \& pre-trained StyleGAN \cite{karras2019style}. \\ \bottomrule
\end{tabular}
\vspace{-0.3cm}
\label{tab:relatedWorks}
\end{table*}

\vspace{0.05cm}
\noindent \textbf{GAN for Vision Tasks:} In a typical GAN model, the generator directly produces new samples from random noise vectors while the discriminator aims to differentiate between real and generator-produced fake samples, improving each other via an adversarial game~\cite{goodfellow2014generative}. With significant progress in design~\cite{brock2018large, karras2017progressive, karras2021alias}, GAN-based methods secured success in a variety of downstream tasks like video generation~\cite{fox2021stylevideogan}, image inpainting~\cite{yu2018generative}, manipulation~\cite{jo2019sc},~\cite{zhu2016generative}, super-resolution~\cite{gabbay2019style}, etc. Generating highly photorealistic outputs, StyleGAN~\cite{karras2019style, karras2020analyzing} introduced a non-linear mapping from input code vector $z \in \mathcal{Z}$ to intermediate latent code $w \in \mathcal{W}$, which controlled the generation process. While traditional GAN is unable to generate conditional output, it can be augmented with additional information to conditionally control the data generation process~\cite{mirza2014conditional}. However, existing conditional generative~\cite{singh2019finegan, li2020mixnmatch} models are unable to inject fine-grained control, especially when conditioned with abstract free-hand sketches.

\noindent \textbf{GAN Inversion:} Exploring GANs has recently led to an interest in inverting a pre-trained GAN~\cite{alaluf2021restyle} for tasks like image manipulation~\cite{alaluf2021restyle}. Typical GAN training aims to learn the weights of generator and discriminator with appropriate loss objectives, to generate random new images $G(z)$ by sampling random noise vectors $z\in\mathcal{Z}$~\cite{goodfellow2014generative}. Contrarily, during GAN inversion, given a reference image we try to find a noise vector $z^*$ in the generator latent space that can accurately reconstruct that image while its weights fixed~\cite{zhu2016generative}. While some methods~\cite{abdal2019image2stylegan, abdal2020image2stylegan++, collins2020editing, creswell2018inverting} directly optimise the latent vector to minimise the reconstruction loss, a few works~\cite{tov2021designing, richardson2021encoding, alaluf2021restyle, guan2020collaborative} train dedicated encoders to find the latent code corresponding to an input image. Among them, optimisation-based methods perform better in terms of reconstruction accuracy, while encoding-based methods work significantly faster. Other methods~\cite{alaluf2022hyperstyle, zhu2020indomain}, take a hybrid approach in order to attain {``the best of both worlds''}~\cite{xia2022gan}. However, images from \textit{different domain} (\textit{e.g.}, semantic label map, edgemap) are not invertible into the latent space of a \textit{photo} pre-trained generator~\cite{bermano2022state}. Consequently, end-to-end trainable methods~\cite{alaluf2021only,chai2021using, nitzan2020disentangling, richardson2021encoding} emerged which aims to map a given image (from a source domain) into the latent space of a pre-trained GAN trained with target domain images. These learned encoders are then used in tasks like, inversion~\cite{alaluf2021restyle}, semantic editing~\cite{alaluf2021restyle}, super-resolution~\cite{menon2020pulse}, face frontalisation~\cite{richardson2021encoding}, inpainting~\cite{richardson2021encoding}.

\vspace{-0.25cm}
\section{Pilot Study: Problems and Analysis}\label{sec:pilot}
\vspace{-0.3cm}

\keypoint{Challenges:}
Sketches being highly abstract in nature, generating a photo from a sketch can have multiple possible outcomes~\cite{richardson2021encoding}.
Generating photorealistic images from sparse sketches incurs three major challenges -- \emph{(i) Locality-bias} assumes that any particular output (\textit{e.g.}, photo) pixel position $I'(x,y)$ is perfectly aligned with the same pixel location $I(x,y)$ of the conditional input (\textit{e.g.}, sketch)~\cite{richardson2021encoding}. However, a free-hand sketch being highly deformed does not necessarily follow the paired photo's intensity boundary~\cite{hertzmann2020line}. \emph{(ii) Hallucinating the colour/texture} in a realistic and contextually-meaningful manner, is difficult from sparse sketch input. \emph{(iii) Deciphering the fine-grained user-intent} is a major bottleneck as the same object can be sketched in diverse ways by different users~\cite{sain2021stylemeup}.

\keypoint{Analysis:}  The popular encoder-decoder architecture~\cite{isola2017image} for converting an input sketch $\mathcal{S}$ to output RGB photo $\mathcal{R}$ via image-to-image translation~\cite{isola2017image} can be formulated as:

\vspace{-0.35cm}
\begin{equation}
    P(\mathcal{R}|\mathcal{S}) = \underbrace{P(\mathcal{Z}|\mathcal{S})}_{\rm Encoder}\underbrace{P(\mathcal{R}|\mathcal{Z})}_{\rm Decoder}
\end{equation}
\vspace{-0.3cm}

\noindent where the encoder $P(\mathcal{Z}|\mathcal{S})$ embeds the sketch into a latent feature $\mathcal{Z}$, from which the decoder $P(\mathcal{R}|\mathcal{Z})$ generates the output photo. Existing works~\cite{huang2018multimodal, choi2020stargan, zhu2017unpaired} have evolved through designing task-specific encoder/decoder frameworks. Despite achieving remarkable success in other image translation problems (\textit{e.g.}, image-restoration~\cite{qu2019enhanced}, colourisation~\cite{isola2017image}), adapting them off-the-shelf fails for our setup. Importantly, we realise that as the loss backpropagates from the decoder to encoder's end, while training~\cite{isola2017image} with sketch-photo pairs $(\mathcal{S},\mathcal{R})$, it implicitly enforces the model to follow sketch as a pseudo edge-boundary~\cite{liu2020unsupervised}. Consequently, the model is \textit{hard-conditioned} by the sketch to treat its strokes as the intensity boundary of the generated photo, thus resulting in a deformed output.

\begin{figure}[!htbp]
    \centering
    \includegraphics[width=\linewidth]{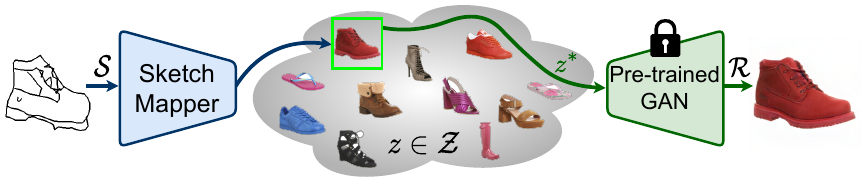}
    \vspace{-0.6cm}
    \caption{{The sketch mapper aims to predict the corresponding latent code of associated photo in the manifold of pre-trained GAN.}}
    \label{fig:solution}
    \vspace{-0.6cm}
\end{figure}

Instead of \textit{end-to-end} encoder-decoder training, we adopt a two stage-approach (\cref{fig:solution}). In the first stage, we model $P(\mathcal{R}|\mathcal{Z})$ as an unsupervised GAN~\cite{goodfellow2014generative}, which being trained from a large number of unlabelled photos of a particular class, is capable of generating realistic photos $G(z)$, given a random vector $z\sim\mathcal{N}(0,1)$~\cite{goodfellow2014generative}. As GAN models learn data distribution~\cite{radford2015unsupervised}, we can loosely assume that any photo can be generated by sampling a \textit{specific} $z^*$ from the GAN latent space~\cite{abdal2019image2stylegan}. Once the GAN model is trained, in the second stage, keeping the $G(\cdot)$ \textit{fixed}, we aim to learn $P(\mathcal{Z}|\mathcal{S})$ as a \emph{sketch mapper} that would encode the input sketch $\mathcal{S}$ into a latent code $\mathcal{Z}$ corresponding to the paired photo $\mathcal{R}$ in the pre-trained GAN latent space.

Advantages of decoupling the encoder-decoder training are twofold -- \textit{(i)} the GAN model~\cite{karras2019style, karras2020analyzing} pre-trained on real photos is bound to generate realistic photos devoid of unwanted deformation, \textit{(ii)} while output quality and diversity of \textit{coupled} encoder-decoder models were limited by the training sketch-photo pairs~\cite{isola2017image}, our \textit{decoupled} decoder being independent of such pairs, can model the large variation of a particular dataset using unlabelled photos~\cite{karras2019style, karras2020analyzing} only.
Sketch-photo pairs are used to train the \emph{sketch mapper} only.

\vspace{-0.3cm}

\section{Background: StyleGAN}
\vspace{-0.1cm}
\noindent In a GAN~\cite{goodfellow2014generative} framework, a generator $G(\cdot)$ aims to generate an image $G(z)$ from a noise vector $z\in\mathcal{Z}$ of size $\mathbb{R}^d$ sampled from a Gaussian distribution~\cite{goodfellow2014generative}, while a discriminator $D(\cdot)$ tries to distinguish between a real and a generated fake image~\cite{goodfellow2014generative}. The training progresses through a two-player minimax game,  thus gradually improving each other over the value function $V(D,G)$ as~\cite{goodfellow2014generative}:
\vspace{-0.2cm}
\begin{equation}
\begin{split}
    \min_{G} \max_{D} V(D, G) = & \ \mathbb{E}_{x \sim p_{data}({x})} [\log D(x)] \\[-5pt]
    & \hspace{-1em} + \mathbb{E}_{z \sim p({z})}[\log(1 - D(G({z})))]
\end{split}
\end{equation}
\vspace{-0.35cm}

Instead of passing a random noise vector $z\in\mathcal{Z}$ directly as the network input~\cite{radford2015unsupervised}, StyleGAN~\cite{karras2019style, karras2020analyzing} eliminates the idea of input layer and always starts from a learned constant tensor of size $\mathbb{R}^{4\times 4\times d}$. The generator network $\mathcal{G}(\cdot)$ consists of a number of progressive resolution blocks, each having the sequence \texttt{conv3$\times$3} $\shortrightarrow$ \texttt{AdaIN} $\shortrightarrow$ \texttt{conv3$\times$3} $\shortrightarrow$ \texttt{AdaIN}~\cite{karras2019style, karras2020analyzing}. StyleGAN employs a non-linear mapping network $f:\mathcal{Z}\rightarrow\mathcal{W}$ (an $8$-layer MLP) to transform $z$ into an intermediate latent vector $w\in\mathcal{W}$ of size $\mathbb{R}^{d}$~\cite{karras2019style, karras2020analyzing}. The \textit{same} latent vector $w$, upon repeatedly passing through a common affine transformation layer $A$ at each level of the generator network, generates the style $y$ = $(y_s, y_b)$ = $A(w)$ for that level. Adaptive instance normalisation (AdaIN)~\cite{karras2019style, karras2020analyzing} is then controlled by $y$ via modulating the feature-map $x_f$ as $\texttt{AdaIN}(x_f,y)=y_{s}\frac{x_f-\mu(x_f)}{\sigma(x_f)}+y_{b}$ after each \texttt{conv3$\times$3} block of $\mathcal{G}(\cdot)$~\cite{karras2019style, karras2020analyzing}. Moreover, stochasticity is injected by adding one-channel uncorrelated Gaussian noise image (per-channel scaled with a learned scaling factor $B$) to each layer of the network before every \texttt{AdaIN} operation~\cite{karras2019style, karras2020analyzing}.

However, due to the limited representability and disentanglement of a \textit{single} latent vector $w\in\mathcal{W}$~\cite{abdal2019image2stylegan}, we embed the input in the extended $\mathcal{W}^+$ latent space~\cite{abdal2019image2stylegan} consisting of \textit{different} latent vectors $w^+\in\mathcal{W^+}$ of size $\mathbb{R}^{k\times d}$, one for each level of the generator network $\mathcal{G}(\cdot)$. For an output of resolution $M\times M$, $k = 2\log_2(M)-2$~\cite{karras2020analyzing}. In this work,  we set $M=256$, making $k=14$.

\vspace{-0.2cm}
\section{Sketch-to-Photo Generation Model}
\vspace{-0.2cm}
\keypoint{Overview:} We aim to devise a sketch-to-photo generation model utilising the rich latent space of a pre-trained StyleGAN~\cite{karras2020analyzing} trained on a particular class to achieve \textit{fine-grained} generation. Once the StyleGAN~\cite{karras2020analyzing} is trained, we fix its weights and train a sketch mapper $\mathcal{E}_s$ that projects an input sketch ($s$) into a latent code $ w_s^+=\mathcal{E}_s(s) \in \mathbb{R}^{14\times d}$  lying in the manifold of pre-trained StyleGAN. In other words, given a sketch input we aim to pick the corresponding latent which when passed through the frozen generator $\mathcal{G}(\cdot)$ would generate an output ($\hat{r}$) most similar to the ground-truth paired photo ($r$).  In particular, we have three salient design components: \textit{(i)} an autoregressive sketch mapper \textit{(ii)} fine-grained discriminative loss besides usual reconstruction objective, and \textit{(iii)} a photo-to-photo mapper $\mathcal{E}_r$ working as a teacher~\cite{furlanello2018born} to improve the learning of $\mathcal{E}_s$.

\vspace{-0.1cm}
\subsection{Model Architecture}\label{sec:5_1}
\vspace{-0.2cm}
\keypoint{Baseline Sketch Mapper:} Inspired by GAN inversion literature~\cite{xia2022gan}, we design our baseline sketch mapper using a ResNet50~\cite{he2016deep} backbone extracting feature map $f_s=\mathcal{F}_s(s) \in \mathbb{R}^{h_f\times w_f\times d}$. In order to generate the latent code of size $\mathbb{R}^{14\times d}$, we use $14$ individual (\emph{not shared}) latent embedding networks (successive stride-two conv-layers with LeakyReLU~\cite{xu2015empirical}), each of them takes $f_s$ as input to predict a $d$-dimensional latent vector. Concatenating them results in the $\mathbb{R}^{14\times d}$ latent code~\cite{richardson2021encoding}. Finally, this latent code upon passing through the pre-trained generator $\mathcal{G}(\cdot)$ generates the output photo $\Hat{r}$. Trained with weighted summation of pixel loss ($l_2$) and perceptual~\cite{zhang2018unreasonable} loss, baseline sketch mapper eventually learns to map an input sketch to its corresponding photo in the latent space of a pre-trained StyleGAN~\cite{karras2020analyzing}.

However, it has a few limitations: Firstly, this baseline ignores the varying levels of sketch abstraction~\cite{yang2021sketchaa}. Ideally, for highly abstract/partial sketches, the output space should be large with many plausible RGB photos~\cite{bhunia2020sketch},
whereas, for a detailed sketch, it should reflect the fine-grained details. Secondly, reconstruction loss \cite{richardson2021encoding} alone, fails to decipher the fine-grained semantic intent of the user.

\keypoint{Autoregressive Latent Mapper:} Instead of predicting the latent code $w_s^+ = \{w_1^+, \cdots, w_k^+ \}$ in one shot, we aim to model it in an \textit{autorgressive} setting keeping a \textit{sequential dependency} among them. Given an input sketch ($s$), the autoregressive sketch mapper $\mathcal{E}_s$ modelling the distribution $P(w_s^+|s)$ can be mathematically expressed as:

\vspace{-0.65cm}
\begin{equation}
P(w_s^+|s) = P(w_1^+, \cdots, w_k^+|s) =   \prod_{i=1}^{k} P(w_i^+ | w^+_{<k}, s)
\end{equation}
\vspace{-0.45cm}

\noindent where the value of the $i^{th}$ predicted latent vector $w_i^+$ depends on all preceding latents. The motivations behind autoregressive modelling are: \textit{(i)} the  disentangled latent space of a StyleGAN depicts semantic feature hierarchy~\cite{yang2021semantic}, where the latent code $w_1^+$ to $w_{14}^+$ controls \textit{coarse} to \textit{fine}-level features. \textit{(ii)} a highly abstract/partial sketch should ideally influence the first few latent vectors governing the major semantic structure, while the later vectors could be sampled randomly from a Gaussian distribution to account for the uncertainty involving such sparse sketches. Whereas, a highly detailed sketch should influence more latent vectors to faithfully reproduce the user's intent.
\textit{(iii)} we aim to synergise the disentangled property of StyleGAN's latent space and the varying levels of sketch abstraction such that the user has the provision to
decide how far should the generated output be \textit{conditioned} on the input sketch and to what extent can it be \textit{hallucinated}.
This is decided by the number of steps of unrolling in the autoregressive process and additionally keeping the later latent codes as random vectors to facilitate multi-modal generation~\cite{richardson2021encoding}.

Given the extracted feature map $f_s = \mathcal{F}_s(s)$, the global average pooled holistic visual feature vector $v_h$ is transformed via a fully-connected (FC) layer to initialise the first hidden state of sequential decoder $f_{seq}(\cdot, \cdot)$ as $h_{0} = \texttt{tanh}(W_h \otimes v_h + b_h)$, with $W_h$, $b_h$ being trainable parameters. At every $j^{th}$ time step, we apply a shared FC-layer on the hidden state $h_j$ to obtain the $j^{th}$ latent code as $w_{k}^+ = W_o \otimes h_j  + b_o$. The current hidden state is updated by $h_k = f_{seq}(h_{k-1}; \eta(f_s, w_{k-1}^{+}))$, where the  previous hidden state of sequential decoder $h_{k-1}$ holds the knowledge about previously predicted latent codes, and $\eta$ models the influence of the formerly predicted latent code on the next prediction along with extracting the relevant information from the feature map $f_s$. In order to model the interaction between $w_k^+$ and $f_s$, we use a simple Hadamard product $\hat{f}_s = f_s  \odot  w_{k-1}^{+} \in \mathbb{R}^{h_f\times w_f\times d}$ which upon passing through successive two-strided convolutional layers followed by LeakyReLU produces a $d$-dimensional output of $\eta$. $f_{seq}(\cdot, \cdot)$ can be modelled using any sequential network (\textit{e.g.}, LSTM~\cite{hochreiter1997long}, RNN~\cite{mcculloch1943logical}, GRU~\cite{cho2014properties}) or self-attention based transformer~\cite{vaswani2017attention} network.
However, here we use GRU~\cite{cho2014properties}, as it was empirically found to be easily optimisable and cheaper while producing compelling results. We wrap this entire process inside the sketch mapper $\mathcal{E}_s$.

To allow multi-modal generation, we always predict a maximum $10$ out of the $14$ unique latent vectors and sample the rest $4$ from Gaussian distribution to inject output variation~\cite{richardson2021encoding}. Moreover, to enforce our model in learning to generate from partial sketches, we introduce a smart \textit{augmentation strategy}, where, we partially render the sketch from $30$-$100\%$ at an interval of $10\%$. While feeding the $\{30\%, 40\%, \cdots 100\%\}$ sketches, we enforce the mapper to predict only the first $m=\{3,4,\cdots 10\}$ corresponding latent vectors. In every case, we pass \textit{random vectors} sampled from Gaussian distribution in place of the remaining ($14$-$m$) \textit{unpredicted vectors}. This strategy ensures that our model eventually learns to generate plausible photos at varying levels of completion, thus allowing the user to control the extent of abstraction as per his/her choice.

\begin{figure}[!tbp]
    \centering
    \includegraphics[width=\linewidth]{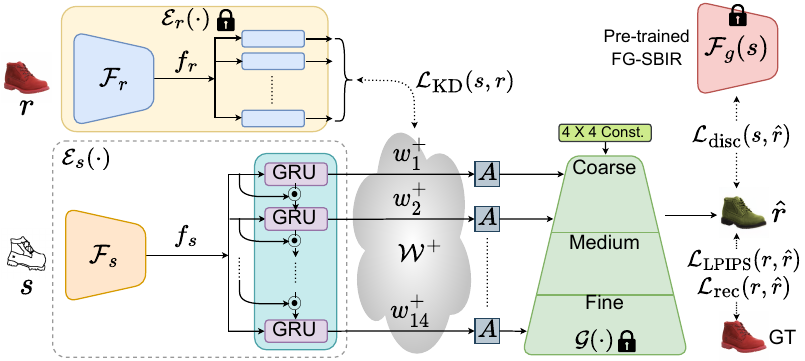}
    \vspace{-0.8cm}
    \caption{$\mathcal{E}_s$ learns to map a \textit{sketch} to the \textit{latent code of its paired photo} in a pre-trained StyleGAN manifold, trained with a mix of reconstruction, fine-grained discriminative, and distillation losses.}
    \label{fig:architecture}
    \vspace{-0.5cm}
\end{figure}

\vspace{-0.1cm}
\subsection{Training Procedure}
\vspace{-0.3cm}
\keypoint{Reconstruction Loss:} Given an input sketch-photo pair $\{s,r\}$ and the generated output photo $\hat{r} = \mathcal{G}(\mathcal{E}_s(s))$, we compute pixel level $l_2$ reconstruction loss as:
\vspace{-0.2cm}
\begin{equation}
\label{eq:l2_loss}
    \mathcal{L}_{\text{rec}}(r,\hat{r})=||r-\hat{r}||_2
\end{equation}
\vspace{-0.6cm}

Besides pixel-wise similarity, we also learn perceptual similarities via LPIPS~\cite{zhang2018unreasonable} loss, which has been found~\cite{guan2020collaborative} to retain photorealism. With $\phi(\cdot)$ as the pre-trained perceptual feature encoder~\cite{zhang2018unreasonable}, LPIPS loss becomes:

\vspace{-0.3cm}
\begin{equation}
\label{eq:lpips_loss}
    \mathcal{L}_{\text{LPIPS}}(r,\hat{r})=||\phi(r)-\phi(\Hat{r})||_2
\end{equation}
\vspace{-0.6cm}

\keypoint{Fine-Grained Discriminative Loss:} While reconstruction loss aims to align the pixel values between generated and ground-truth photo, the \textit{discriminative} sketch-photo (paired) association compared to other photos needs to be modelled further to reflect the \textit{fine-grained user intent} of input sketch in the output space. Triplet with cosine-distance based pre-trained fine-grained SBIR~\cite{chowdhury2022partially} model $\mathcal{F}_g (\cdot)$ places a sketch nearer to its \textit{paired} photo compared to others in a joint-embedding space. Therefore, we compute a discriminative fine-grained loss that measures the cosine similarity between $s$ and $\Hat{r}$ as:

\vspace{-0.45cm}
\begin{equation}
\label{eq:fgsbir_ot_loss}
    \mathcal{L}_{\text{disc}}(s,\Hat{r}) = 1 - \frac{\mathcal{F}_g(s) \cdot \mathcal{F}_g(\Hat{r})}{||\mathcal{F}_g(s)||~~||\mathcal{F}_g(\Hat{r})||}
    \vspace{-0.15cm}
\end{equation}

\keypoint{Photo-to-Photo Mapper as Teacher:}
Photo-to-photo mapping being an easier task than sketch-to-photo underpins our motivation towards introducing a photo-to-photo mapper $\mathcal{E}_r(\cdot)$ as a teacher~\cite{furlanello2018born} to additionally guide the learning of our sketch-mapper~$\mathcal{E}_s(\cdot)$, thus handling the subjective nature of sketches and its resultant large sketch-photo domain gap. Architecturally, $\mathcal{E}_r(\cdot)$ is identical to our baseline $\mathcal{E}_s(\cdot)$ with the aim of reconstructing the input photo ($r$) at the output ($\Hat{r}$): $\mathcal{G}(\mathcal{E}_r(r)) \approx \Hat{r}$. Once trained, latent vectors predicted by $\mathcal{E}_r$ (weights frozen) acts as a ground-truth additionally supervising $\mathcal{E}_s$ via a distillation loss as:

\vspace{-0.4cm}
\begin{equation}
    \mathcal{L}_{\text{KD}}(s,r) = ||\mathcal{E}_s(s)-\mathcal{E}_r(r)||_2
\end{equation}
\vspace{-0.5cm}

\noindent We impose $\mathcal{L}_{\text{KD}}$ only on the predicted latents (max $10$) \textit{not} on the random ones. Our overall training objective is $\mathcal{L}_{\text{total}}=\lambda_{1}\mathcal{L}_{\text{rec}} + \lambda_{2}\mathcal{L}_{\text{LPIPS}} + \lambda_{3}\mathcal{L}_{\text{disc}} + \lambda_{4}\mathcal{L}_{\text{KD}}$.

\vspace{-0.2cm}
\section{Experiments}
\vspace{-0.25cm}
\keypoint{Dataset:} UT Zappos50K~\cite{yu2014fine} and pix2pix Handbag~\cite{isola2017image} datasets are used to pre-train the StyleGAN generator in \textit{shoe} and \textit{handbag} classes respectively. While for \textit{chair} class, we collected over $10,000$ photos from websites like IKEA, ARGOS, etc., we used QMUL-ShoeV2~\cite{bhunia2020sketch, song2018learning}, QMUL-ChairV2~\cite{bhunia2020sketch, song2018learning}, and Handbag~\cite{song2017deep} datasets containing sketch-photo pairs to train the sketch mapper. Out of $6730$/$1800$/$568$ sketches and $2000$/$400$/$568$ photos from ShoeV2/ChairV2/Handbag datasets, $6051$/$1275$/$400$ sketches and $1800$/$300$/$400$ photos are used for training respectively, keeping the rest for testing. Notably, StyleGAN pre-training \textit{does not} involve any sketch-photo pairs.

\keypoint{Implementation Details:} Adam~\cite{kingma2014adam} optimiser is used to pre-train a category specific StyleGAN~\cite{karras2020analyzing} with feature embedding size of $d=512$ for $8M$ iterations at learning rate of $10^{-3}$ and batch size $8$. {Based on empirical observations, we disable path-length regularisation \cite{karras2020analyzing} and reduce $R1$ regularisation's weight to $2$ for superior quality and diversity.} We use a combination of Rectified Adam~\cite{liu2019variance} and Lookahead~\cite{zhang2019lookahead} method as an optimiser to train the sketch-to-photo mapper for $5M$ iterations at a constant learning rate of $10^{-5}$ and a batch size of $4$. $\lambda_{1}$, $\lambda_{2}$, $\lambda_{3}$, and $\lambda_{4}$ are set to $1$, $0.8$, $0.5$, and $0.6$ respectively.

\keypoint{Evaluation:} We use four metrics -- \textit{(i) Fr\'echet Inception Distance (FID)}~\cite{karras2019style}: uses pre-trained InceptionV3's activation distribution statistics to estimate distance between synthetic and real data where a lower value indicates better generation quality. \textit{(ii) Learned Perceptual Image Patch Similarity (LPIPS)}~\cite{zhang2018unreasonable}: is a weighted $l_2$ distance between two ImageNet-pretrained AlexNet~\cite{krizhevsky2017imagenet}-extracted deep features of ground-truth and generated images. A higher LPIPS value denotes better diversity. \textit{(iii) Mean Opinion Score (MOS)}: for human study, each of the $30$ human workers was asked to draw $50$ sketches in our system, and rate every generated photo (both ours and competitor's) on a scale of $1$ to $5$~\cite{huynh2010study} (bad$\rightarrow$excellent) based on their \textit{opinion} of how closely it matched their \textit{photorealistic imagination} of the associated sketch. For each method, we compute the final MOS value by taking the mean ($\mu$) and variance ($\sigma$) of all $1500$ of its MOS responses. \textit{(iv) Fine-Grained Metric (FGM)}: to judge the \textit{fine-grainedness} of sketch mapping, we propose a new metric, which uses features from a pre-trained FG-SBIR model~\cite{yu2016sketch} to compute cosine similarity between input sketch and generated photo. A higher FGM value denotes better fine-grained association between them.

\keypoint{Competitors:} We compare our proposed framework with various state-of-the-art (SOTA) methods and two self-designed baselines. Among those, \textbf{pix2pix}~\cite{isola2017image} uses a conditional generative model for sketch-to-photo translation. \textbf{MUNIT}~\cite{huang2018multimodal} aims to produce diverse outputs given one input sketch. It tries to decompose an image into a \textit{content} and a \textit{style} code followed by learning those codes simultaneously. \textbf{CycleGAN}~\cite{zhu2017unpaired} utilises cycle-consistency loss with a GAN model for bidirectional image-to-image translation. \textbf{U-GAT-IT}~\cite{kim2019u} uses an attention module for image translation while focusing on the domain-discriminative parts. Moreover, employing a pre-trained StyleGAN~\cite{karras2020analyzing} we compare with the baseline \textbf{B-Sketch Mapper} which is equivalent to the baseline sketch mapper described in \cref{sec:5_1}. Following optimisation-based GAN inversion~\cite{abdal2019image2stylegan}, we design \textbf{B-Sketch Optimiser} where we iteratively optimise the latent code using input sketch as a ground-truth with perceptual loss~\cite{zhang2018unreasonable}. For a fair comparison, we trained all competing methods in a supervised manner with sketch-photo pairs from ShoeV2, ChairV2, and Handbag datasets.

\vspace{-0.10cm}
\subsection{Performance Analysis \& Discussion} \label{sec:6_1}
\vspace{-0.25 cm}
\begin{figure*}[!htbp]
    \centering
    \includegraphics[width=0.95\textwidth]{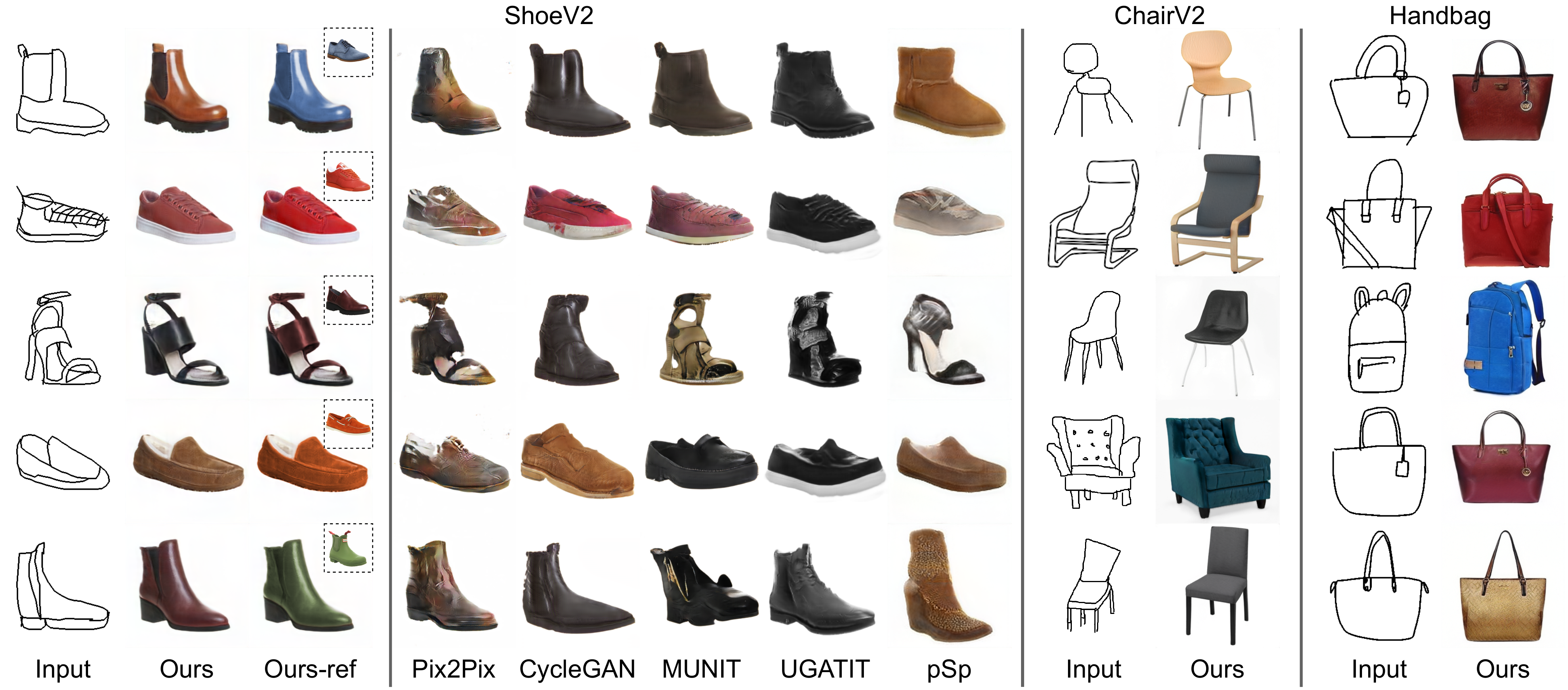}
    \vspace{-0.3cm}
    \caption{Qualitative comparison with various state-of-the-art competitors on ShoeV2 dataset. \textbf{Ours-ref} (column 3) results depict that our method can faithfully replicate the appearance of a given reference photo (shown in the top-right inset).}
    \label{fig:qualitative}
    \vspace{-0.20cm}
\end{figure*}

\begin{table*}[!htbp]
\renewcommand{\arraystretch}{0.7}
\setlength{\tabcolsep}{5pt}
\footnotesize
\centering
\caption{Benchmarks on ChairV2, ShoeV2, and Handbag datasets.}
\vspace{-0.3cm}
\label{tab:benchmark}
\begin{tabular}{lcccccccccccc}
\toprule
\multirow{3}{*}{Methods} & \multicolumn{4}{c}{ChairV2} & \multicolumn{4}{c}{ShoeV2} & \multicolumn{4}{c}{Handbag} \\ \cmidrule(lr){2-5}\cmidrule(lr){6-9}\cmidrule(lr){10-13}
& \multirow{2}{*}{FID$\downarrow$}  & \multirow{2}{*}{LPIPS$\uparrow$}  & MOS$\uparrow$ & \multirow{2}{*}{FGM$\uparrow$} & \multirow{2}{*}{FID$\downarrow$}    & \multirow{2}{*}{LPIPS$\uparrow$}  & MOS$\uparrow$ & \multirow{2}{*}{FGM$\uparrow$} & \multirow{2}{*}{FID$\downarrow$}    & \multirow{2}{*}{LPIPS$\uparrow$}  & MOS$\uparrow$ & \multirow{2}{*}{FGM$\uparrow$} \\
&   &   & $\mu\pm\sigma$  & &   &  &  $\mu\pm\sigma$  & &   &   & $\mu\pm\sigma$ &\\
\cmidrule(lr){1-5}\cmidrule(lr){6-9}\cmidrule(lr){10-13}
pix2pix~\cite{isola2017image}       & 177.79 & 0.096 & 2.32$\pm$0.7 & 0.51 & 65.09 & 0.071 & 2.11$\pm$0.1& 0.58 & 184.57 & 0.074 & 2.94$\pm$0.3 & 0.41\\
MUNIT~\cite{huang2018multimodal}    & 168.81 & 0.264 & 2.28$\pm$0.3 & 0.37 & 92.21 & 0.248 & 2.01$\pm$0.5& 0.49 & 175.68 & 0.163 & 2.11$\pm$0.2 & 0.33\\
CycleGAN~\cite{zhu2017unpaired}     & 124.96 & 0.000 & 2.38$\pm$0.1 & 0.45 & 79.35 & 0.000 & 2.64$\pm$0.6 & 0.53 & 150.11 & 0.000 & 2.87$\pm$0.1 & 0.38\\
U-GAT-IT~\cite{kim2019u}            & 107.24 & 0.000 & 2.71$\pm$0.8 & 0.32 & 76.89 & 0.000 & 2.87$\pm$0.7 & 0.44 & 127.49 & 0.000 & 2.96$\pm$0.5 & 0.30\\
pSp~\cite{richardson2021encoding}   & 105.54 & 0.325 & 3.64$\pm$0.1 & 0.60 & 54.48 & 0.298 & 3.01$\pm$0.9 & 0.67 & 122.54 & 0.298 & 3.52$\pm$0.7 & 0.51\\
\cmidrule(lr){1-5}\cmidrule(lr){6-9}\cmidrule(lr){10-13}
B-Sketch Optimiser                  & 138.40 & 0.135 & 2.15$\pm$0.6 & 0.28 & 63.52 & 0.127 & 2.08$\pm$0.1 & 0.31 & 163.32 & 0.104 & 2.17$\pm$0.2 & 0.24\\
B-Sketch Mapper                     & 111.99 & 0.228 & 3.51$\pm$0.3 & 0.56 & 57.27 & 0.218 & 3.14$\pm$0.2 & 0.61 & 130.87 & 0.138 & 3.01$\pm$0.3 & 0.45\\
\rowcolor{Gray}
Proposed                            & \bf90.21 & \bf0.507 & \bf4.69$\pm$0.1 & \bf0.79 & \bf35.85 & \bf0.489 & \bf4.24$\pm$0.5 & \bf0.88 & \bf100.23 & \bf0.408 & \bf4.16$\pm$0.1 & \bf0.72 \\ \bottomrule
\end{tabular}
\vspace{-0.4cm}
\end{table*}

\keypoint{Result Analysis:} The proposed method consistently surpasses (\Cref{tab:benchmark}) other state-of-the-arts in terms of quality (FID), and diversity (LPIPS). \textbf{Pix2pix}~\cite{isola2017image} with its naive conditional image-to-image translation formulation is outperformed by \textbf{CycleGAN}~\cite{zhu2017unpaired} (by $-14.26$ FID on ShoeV2), as the latter is reinforced with a cycle consistency loss in an adversarial training paradigm in addition to the bidirectional guidance. \textbf{U-GAT-IT}~\cite{kim2019u} with its attention-based formulation, surpasses others proving the efficacy of attention-module in image translation tasks. Although \textbf{MUNIT}~\cite{huang2018multimodal} and \textbf{pSp}~\cite{richardson2021encoding} supports multi-modal generation, our method, excels both in terms of output diversity ($0.489$ LPIPS on ShoeV2). Naive Baselines of \textbf{B-Sketch Mapper} and \textbf{B-Sketch Optimiser} with their simplistic design fall short of surpassing the proposed framework. Our method achieves the highest (\Cref{tab:benchmark}) degree of fine-grained association ($0.88$ FGM on ShoeV2), thanks to its novel fine-grained discriminative loss. When compared to our framework, there exists a noticeable deformity in the photos generated by its competitors (\cref{fig:qualitative}). Photos generated by \textbf{pix2pix}~\cite{isola2017image}, \textbf{MUNIT}~\cite{huang2018multimodal} and \textbf{CycleGAN}~\cite{zhu2017unpaired} suffer from deformity and lack of photorealism. Although \textbf{U-GAT-IT}~\cite{kim2019u} and \textbf{pSp}~\cite{richardson2021encoding} outputs are somewhat realistic, they are mostly unfaithful to the input sketch. As observed from \cref{fig:qualitative}, the photos generated by SOTA methods almost invariably fail to capture the \textit{semantic intent} of the user, yielding deformed images. Contrarily, given our visually pleasing (\cref{fig:qualitative}), and richer generation quality, our method vastly outperforms most SOTA and baselines in terms of MOS value (\Cref{tab:benchmark}). Furthermore, our method can replicate the appearance of a given photo onto the generated one (\cref{fig:qualitative}) by predicting coarse and mid-level latent codes from the input sketch and taking the fine-level codes of the reference photo predicted by our photo-to-photo mapper.

In summary, with the help of smooth \cite{bermano2022state} latent space of StyleGAN~\cite{karras2019style, karras2020analyzing} along with auto-regressive sketch mapper and the fine-grained discriminative loss, our approach almost always ensures photorealistic translations with accurate reproduction of users intent in the target domain.

\begin{figure}[!htbp]
    \centering
    \includegraphics[width=\linewidth]{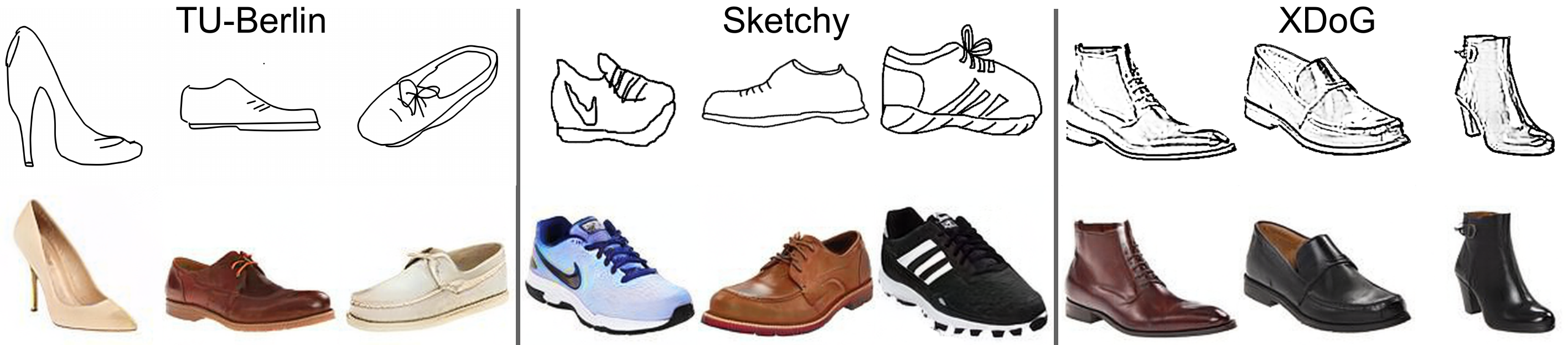}
    \vspace{-0.6cm}
    \caption{Generalisation across sketch styles.}
    \label{fig:generalisation}
\vspace{-0.3cm}
\end{figure}

\vspace{-0.1cm}
\keypoint{Generalisation onto Unseen Dataset:} \cref{fig:generalisation} shows a few shoe sketches randomly sampled from Sketchy~\cite{sangkloy2016sketchy} and TU-Berlin~\cite{eitz2012humans} datasets, and a few XDoG~\cite{winnemoller2012xdog} edgemaps. While the edgemaps are perfectly pixel-aligned, sketches show significant shape deformation and abstraction. However, our model trained on ShoeV2 generalises well to all unseen sketch styles, yielding compelling results.

\vspace{-0.2cm}

\begin{figure}[!htbp]
    \centering
    \includegraphics[width=\linewidth]{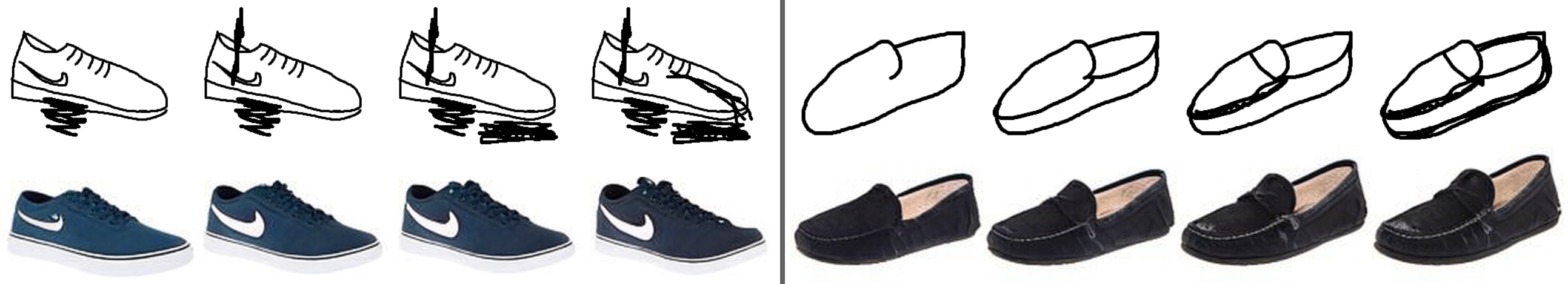}
    \vspace{-0.6cm}
    \caption{Examples showing the effect of noisy stroke addition (left) and generation from partial sketches (right).}
    \label{fig:robust}
    \vspace{-0.2cm}
\end{figure}

\keypoint{Robustness and Sensitivity:} The free-flow style of amateur sketching is likely to introduce irrelevant noisy strokes \cite{bhunia2022sketching}. To prove our model's robustness to noise, during testing, we gradually add synthetic noisy strokes~\cite{liu2020unsupervised} onto clean input sketches. Meanwhile, to assess \textit{sensitivity} to partial sketches \cite{bhunia2020sketch}, we render input sketches partially at $25\%$, $50\%$, $75\%$, and $100\%$ completion-levels before generation. We observe (\cref{fig:robust} (right)) that our method is resilient to partial inputs, and the output quality remains steady even when the input sketches are extremely noisy (\cref{fig:robust} (left)). As our method is not \textit{hard}-conditioned on input sketches, noise-addition or partial-completion has negligible impact on the final output, thus achieving an impressive FID score of $49.6$ even with the addition of $80\%$ noisy strokes.

\vspace{-0.3cm}

\vspace{-0.25cm}
\subsection{Ablation on Design}
\vspace{-0.15cm}
\noindent{\textbf{[i] Benefit of $\mathcal{W^+}$-space embedding:}} To assess the contribution of $\mathcal{W^+}$ latent space embedding, we design two experiments -- \textit{(a)} \textbf{$\mathcal{W}$ latent space}, and \textit{(b)} \textbf{Naive $\mathcal{W^+}$}. For $\mathcal{W}$ latent space, given $s$, we employ a generic ResNet50~\cite{he2016deep} back-boned encoder producing a single latent vector $w\in\mathcal{W}$ of size $\mathbb{R}^{d}$ which upon \textit{repeatedly} passing through every level of StyleGAN, generates an output. Whereas, for Naive $\mathcal{W^+}$ encoding, we extend \textit{(a)} with an additional layer to convert the $w\in\mathcal{W}$ latent vector to $w^+\in\mathcal{W^+}$ latent code of size $\mathbb{R}^{14\times d}$. Despite Naive $\mathcal{W^+}$ achieving lower FID than $\mathcal{W}$ latent embedding, it causes a drastic FID surge ($11.02$ on ShoeV2) when compared to \textbf{Ours-full} model~(\Cref{tab:ablation}). It shows how the proposed method improves output quality and diversity compared to naive embedding in the $\mathcal{W}$ or $\mathcal{W^+}$. \textbf{[ii] Effect of FG-discriminative loss:} Fine-grained discriminative loss aims to minimise the sketch-photo domain gap. Eliminating it causes a stark increase in FID of $14.44$ on ShoeV2 dataset (\Cref{tab:ablation}). We hypothesise that this drop is due to the lack of cross-domain regularisation offered by the fine-grained discriminative loss. Furthermore, as evident from the \textbf{w/o FG-SBIR loss} result in \Cref{tab:ablation}, it apparently provides further guidance for better correlating a sketch-photo pair.  \textbf{[iii] Choice of photo-to-photo mapper as teacher:} Training a good teacher~\cite{furlanello2018born} network should not only be free of additional label-cost but should also be well-suited to the student network's objective~\cite{bhunia2021more}.  We posit that the photo-to-photo task is meaningful in this scenario, as the GAN was pre-trained on photos \textit{only}, without access to any sketch. As seen in \Cref{tab:ablation}, omitting the $\mathcal{E}_r$ teacher network results in a noticeable drop in performance (FID of $11.02$ on ShoeV2, confirming that $\mathcal{E}_r$ as a teacher-assistant handles the large sketch-photo domain gap efficiently. \textbf{[iv] Does autoregressive mapping help?} To judge the contribution of our autoregressive modelling, we replaced the autoregressive module with the baseline latent mapper explained in \cref{sec:5_1}. In \textbf{w/o autoregressive}, we see a significant dip ($21.42$ FID drop in ShoeV2) in the output quality. A probable reason might be that the autoregressive module helps in the sequential unrolling of the abstractness of an input sketch, thus aiding in better semantic understanding.

\vspace{-0.3cm}
\begin{table}[!hbt]
\setlength{\tabcolsep}{7pt}
\renewcommand{\arraystretch}{0.5}
\centering
\caption{{Ablation on design.}}
\vspace{-0.3cm}
\label{tab:ablation}
\footnotesize
\begin{tabular}{lcccc}
\toprule
\multirow{2}{*}{Methods} & \multicolumn{2}{c}{ChairV2} & \multicolumn{2}{c}{ShoeV2} \\\cmidrule(lr){2-3}\cmidrule(lr){4-5}
& FID$\downarrow$ & LPIPS$\uparrow$ & FID$\downarrow$ & LPIPS$\uparrow$ \\
\cmidrule(lr){1-3}\cmidrule(lr){4-5}
w/o autoregressive           & 111.99  & 0.228  & 57.27 & 0.218 \\
w/o FG-SBIR loss             & 104.29  & 0.425  & 50.29 & 0.417 \\
w/o $\mathcal{E}_r$ teacher  & 99.38  & 0.418  & 46.87 & 0.404 \\ \cmidrule(lr){1-3}\cmidrule(lr){4-5}
Naive $\mathcal{W^+}$        & 99.24  & 0.401  & 46.87 & 0.368 \\
$\mathcal{W}$ latent space   & 107.99  & 0.359  & 52.35 & 0.344 \\
\rowcolor{Gray}
Ours-full                    & \bf90.21 & \bf0.507 & \bf35.85 & \bf0.489 \\\bottomrule
\end{tabular}
\vspace{-0.3cm}
\end{table}

\vspace{-0.3cm}
\begin{table}[!htbp]
\setlength{\tabcolsep}{5pt}
\renewcommand{\arraystretch}{0.7}
\centering
\caption{{Results for standard FG-SBIR task.}}
\vspace{-0.3cm}
\label{tab:fgsbir}
\footnotesize
\begin{tabular}{lcccc}
\toprule
\multirow{2}{*}{Methods} & \multicolumn{2}{c}{ChairV2} & \multicolumn{2}{c}{ShoeV2} \\\cmidrule(lr){2-3}\cmidrule(lr){4-5}
& Acc.@1 & Acc.@5 & Acc.@1 & Acc.@5 \\
\cmidrule(lr){1-3}\cmidrule(lr){4-5}
Triplet-SN~\cite{yu2016sketch}          & 47.4  & 71.4  & 28.7 & 63.5 \\
HOLEF-SN~\cite{song2017deep}       & 50.7  & 73.6  & 31.2 & 66.6 \\
StyleMeUp~\cite{sain2021stylemeup}      & 62.8  & 79.6  & 36.4 & 68.1 \\
CrossHier~\cite{sain2020cross}          & 62.8  & 79.1  & 36.2 & 67.8 \\
Semi-Sup~\cite{bhunia2021vectorization} & 60.2  & 78.1  & 39.1 & 69.9 \\
\rowcolor{Gray}
Proposed                                &\bf65.1  &\bf79.2  &\bf44.1  &\bf75.1 \\\bottomrule
\end{tabular}
\end{table}
\vspace{-0.3cm}

\vspace{-0.2cm}
\subsection{Downstream Applications}
\vspace{-0.2cm}
\keypoint{Fine-Grained SBIR:} Fine-grained SBIR aims at retrieving a particular image given a query sketch~\cite{yu2016sketch}. Here, we perform retrieval by first translating a query sketch into the photo domain, and then finding its nearest neighbourhood feature match in the entire photo gallery using an ImageNet pre-trained VGG-16~\cite{simonyan2014very} feature extractor. Hence, we essentially convert the \textit{sketch-based} retrieval task into an \textit{image-based} retrieval task. As seen in \Cref{tab:fgsbir}, our method beats SOTA FG-SBIR schemes~\cite{yu2016sketch, song2017deep, sain2021stylemeup, sain2020cross, bhunia2021vectorization} in terms of Acc.@q, which measures the percentage of sketches having a true-paired photo in the top-q retrieved list.

\begin{figure}[!htbp]
    \centering
    \includegraphics[width=\linewidth]{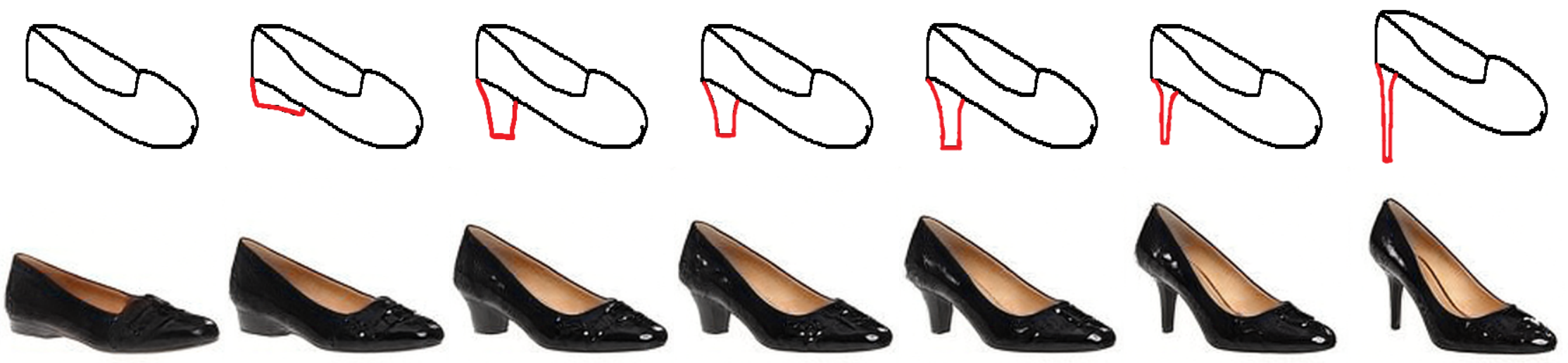}
    \vspace{-0.6cm}
    \caption{Sketches of an editing episode (edited strokes in \red{red}) and corresponding output photos.}
    \label{fig:local_edit}
    \vspace{-0.4cm}
\end{figure}

\keypoint{Precise Semantic Editing:} Local semantic image editing is a popular application of GAN inversion~\cite{alaluf2021restyle}. Our method enables realistic semantic editing, where modifying one region of an input sketch, yields seamless local alterations in the generated images. \cref{fig:local_edit} depicts one such sketch editing episode where the user gradually changes the heel length via sketch, to observe consistent local changes in the output photo domain. To our best knowledge, this is one of the first attempts towards such \textit{fine-grained} semantic editing.

\vspace{-0.3cm}

\begin{figure}[!htbp]
    \centering
    \includegraphics[width=\linewidth]{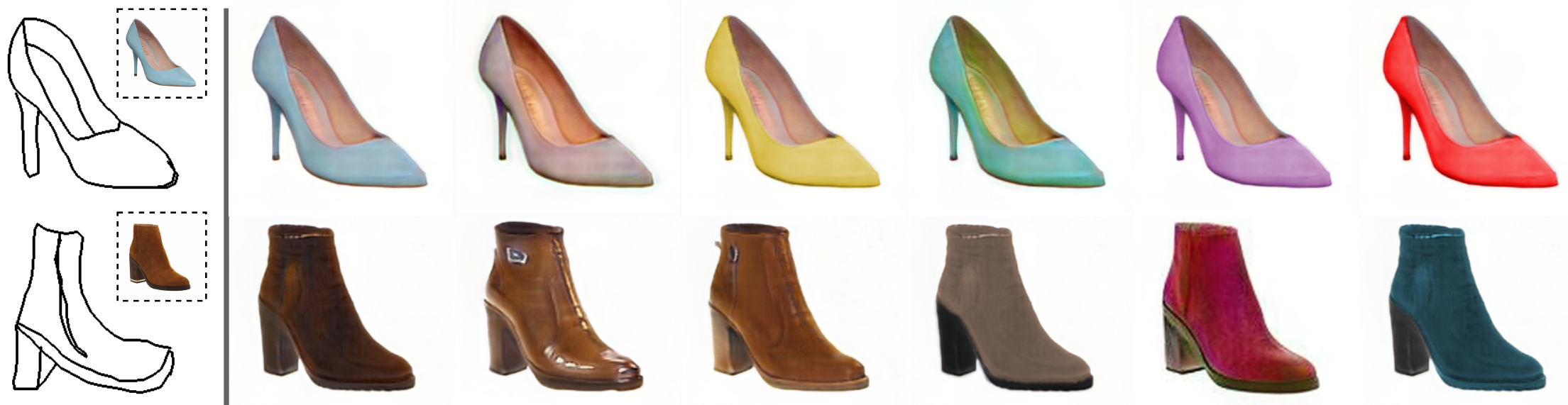}
    \vspace{-0.6cm}
    \caption{Multi-modal generation showing varied colour (top), appearance features (bottom). Reference photo shown in inset.}
    \label{fig:shuffling}
    \vspace{-0.4cm}
\end{figure}

\begin{figure}[!htbp]
    \centering
    \includegraphics[width=\linewidth]{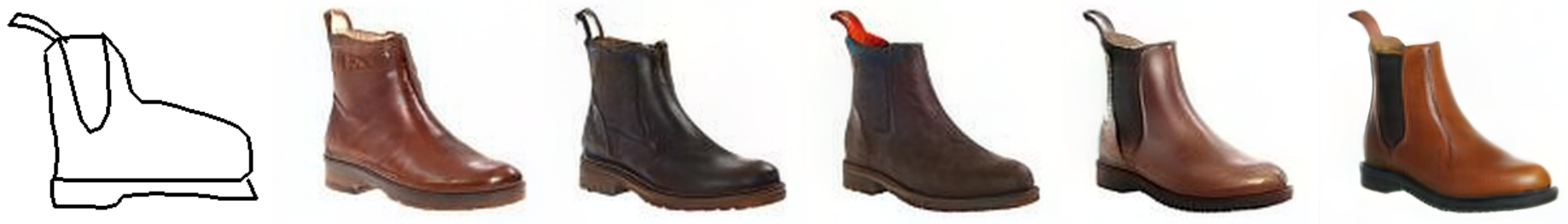}
    \vspace{-0.6cm}
    \caption{(Left to right) Generation by using increasing numbers ($\{2,4,6,8,10\}$) of $d$-dimensional latent vectors.}
    \label{fig:unroll}
    \vspace{-0.3cm}
\end{figure}

\keypoint{Fine-grained Control:} The proposed method also allows multi-modal generation with \textit{fine-grained} appearance control by replacing \cite{richardson2021encoding} medium or fine-level latent codes with random vectors (\cref{fig:shuffling}). Furthermore, \cref{fig:unroll} shows results with increasing number of unrolling (\cref{sec:5_1}) steps, where detail gets added progressively with every increasing step.

\vspace{-0.1cm}

\vspace{-0.1cm}

\vspace{-0.1cm}
\section{Conclusion}
\vspace{-0.1cm}
We address a key challenge for conditional sketch-to-photo generation -- existing models consider input abstract sketches as a \textit{hard constraint}, resulting in deformed output images. A novel supervised sketch-to-photo generation model is proposed that explicitly handles sketch-photo locality bias, enabling it to generate photorealistic images even from highly abstract sketches. It is based on an autoregressive latent mapper, that maps a sketch to a pre-trained StyleGAN's latent space to generate an output. Extensive experiments show our method to outperform existing state-of-the-arts.

{\small
\bibliographystyle{ieee_fullname}
\bibliography{arxiv}
}

\clearpage

\onecolumn{
\centering
\title{\Large{\textbf{Supplementary material for \\ \textit{Picture that Sketch}: Photorealistic Image Generation from Abstract Sketches}}\vspace{0.5cm}}
\author{Subhadeep Koley\textsuperscript{1,2} \hspace{.2cm} Ayan Kumar Bhunia\textsuperscript{1} \hspace{.2cm} Aneeshan Sain\textsuperscript{1,2} \hspace{.2cm} Pinaki Nath Chowdhury\textsuperscript{1,2} \\
Tao Xiang\textsuperscript{1,2}\hspace{.3cm}  Yi-Zhe Song\textsuperscript{1,2} \\
\textsuperscript{1}SketchX, CVSSP, University of Surrey, United Kingdom.  \\
\textsuperscript{2}iFlyTek-Surrey Joint Research Centre on Artificial Intelligence.}\\
\tt\small \{s.koley, a.bhunia, a.sain, p.chowdhury, t.xiang, y.song\}@surrey.ac.uk

\date{}
}

\maketitle
\thispagestyle{empty} 

\section*{A. Additional Results}
\begin{figure}[!htbp]
    \centering
    \includegraphics[width=0.9\linewidth]{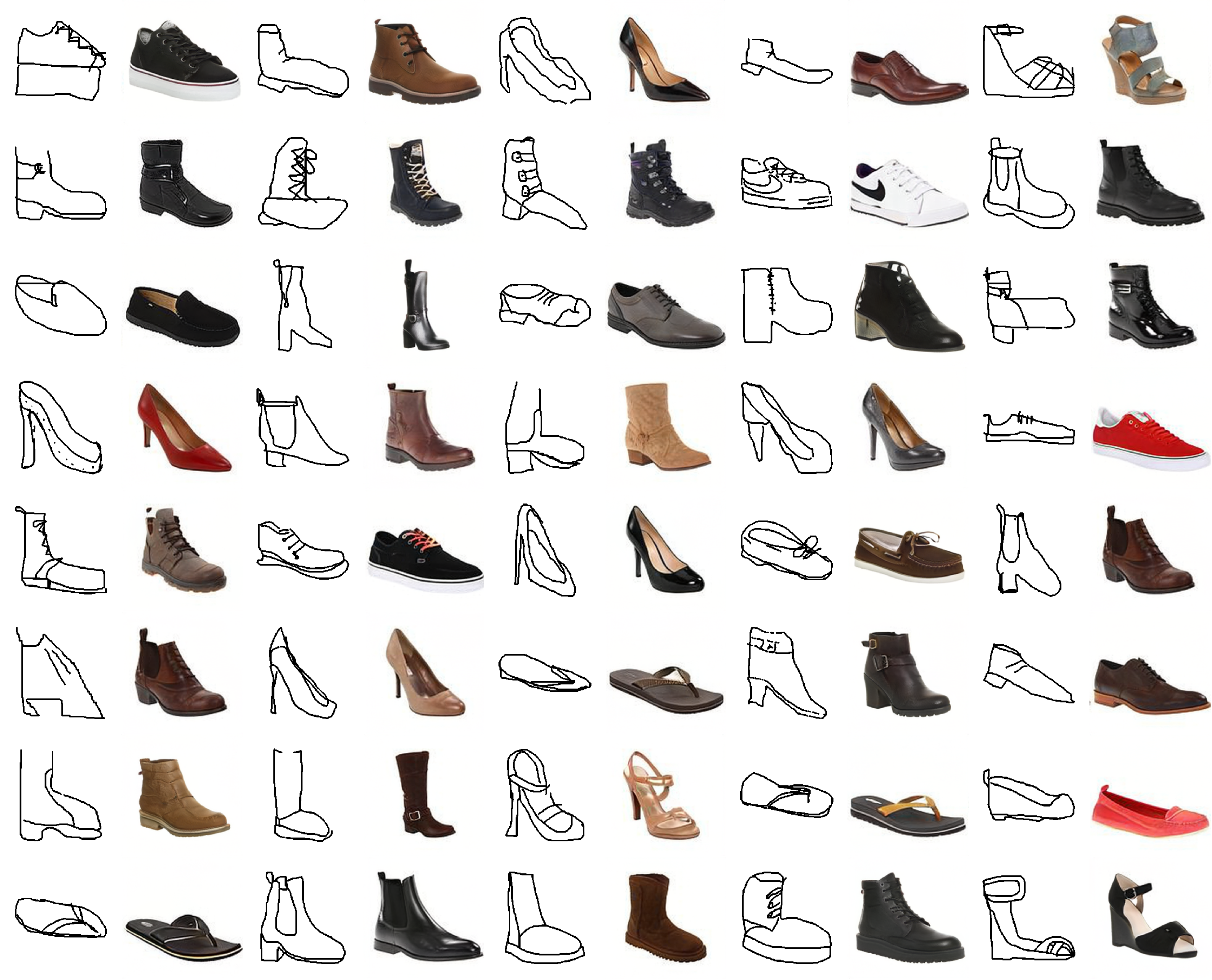}
    \caption{Results generated by the proposed method with sketches from the QMUL-ShoeV2~\cite{bhunia2020sketch, song2018learning} dataset.}
    \label{fig:shoe1}
\end{figure}

\begin{figure}[!htbp]
    \centering
    \includegraphics[width=0.9\linewidth]{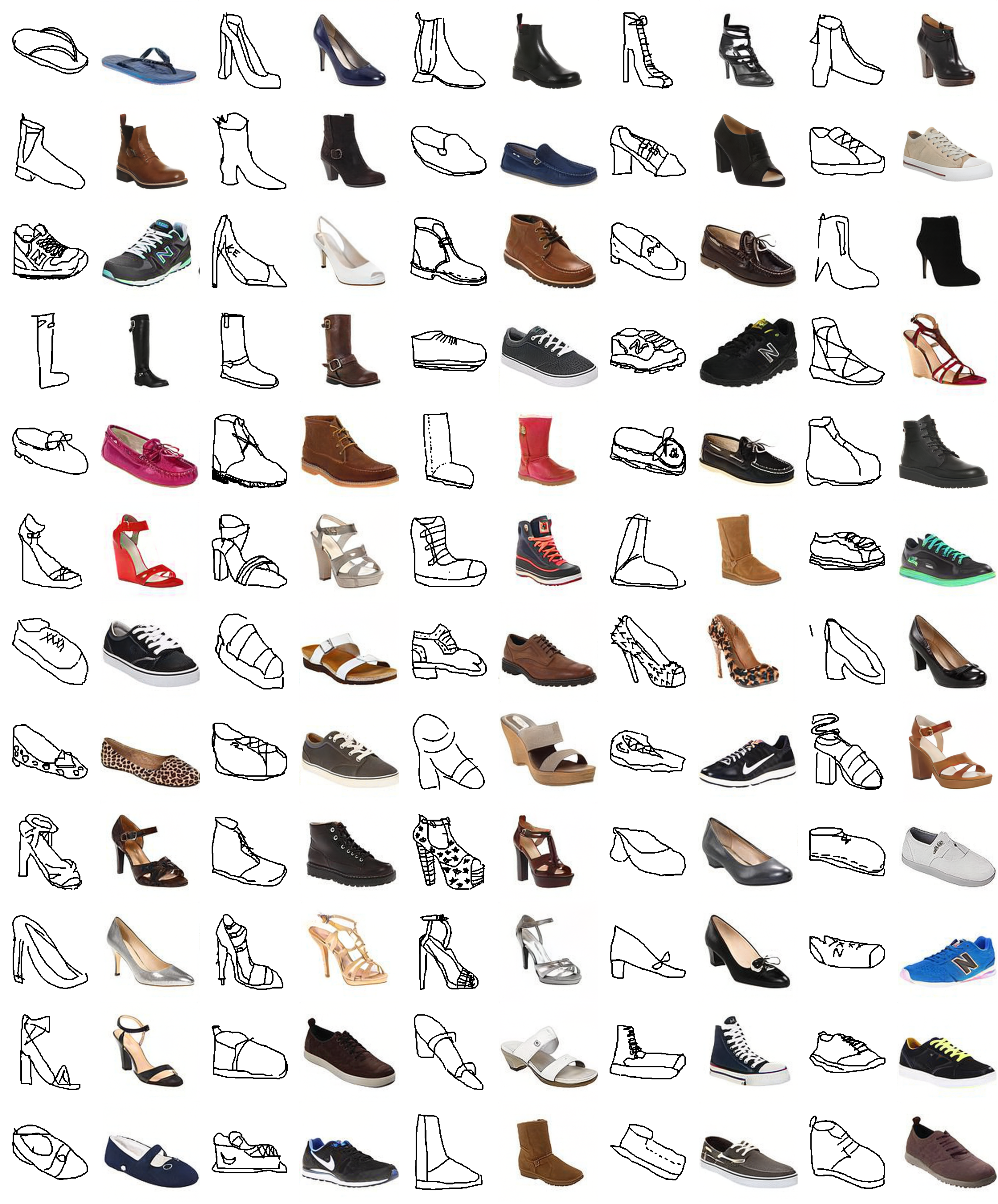}
    \caption{Results generated by the proposed method with sketches from the QMUL-ShoeV2~\cite{bhunia2020sketch, song2018learning}~\cite{bhunia2020sketch, song2018learning} dataset.}
    \label{fig:shoe2}
\end{figure}

\begin{figure}[!htbp]
    \centering
    \includegraphics[width=0.9\linewidth]{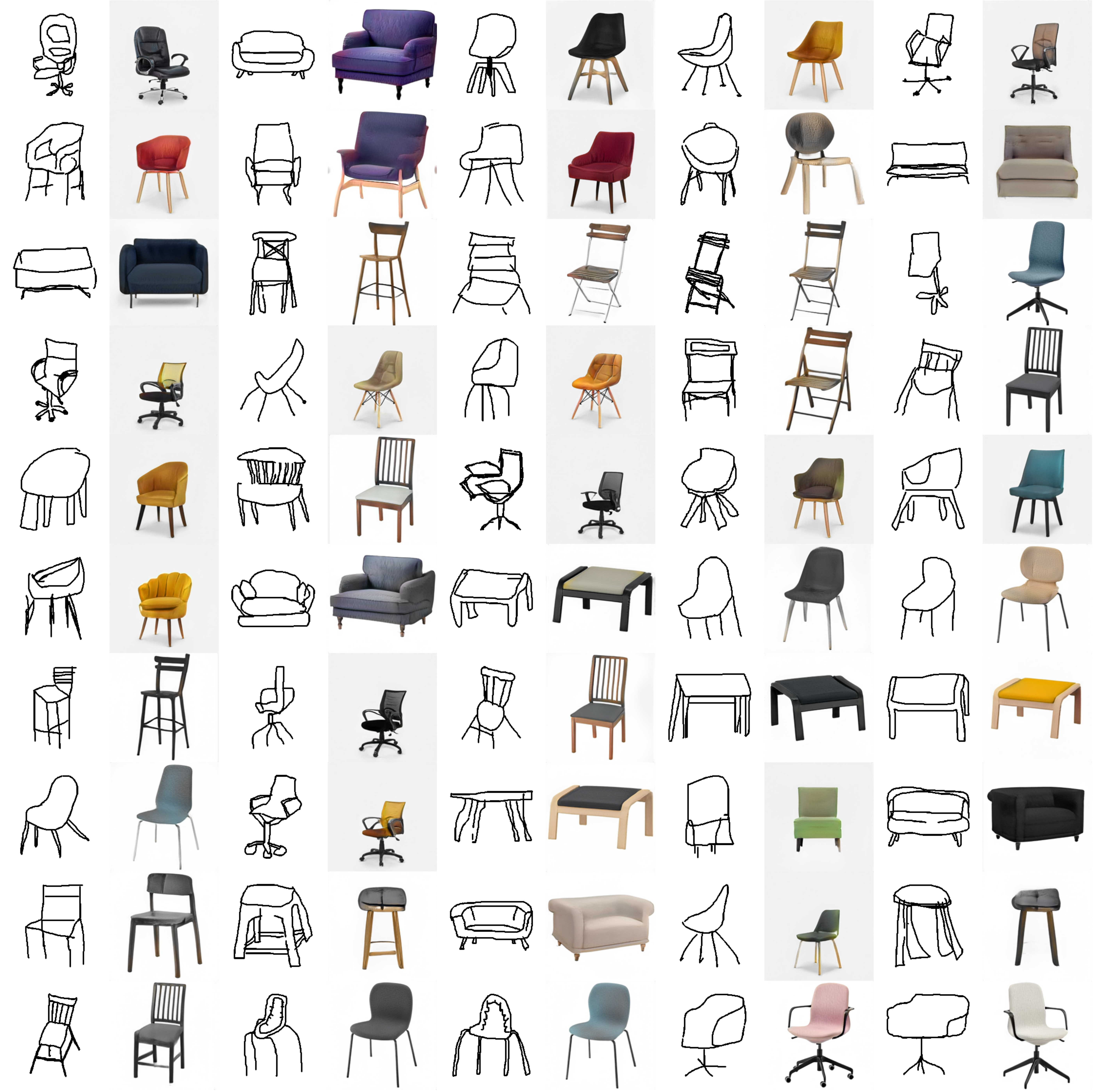}
    \caption{Results generated by the proposed method with sketches from the QMUL-ChairV2~\cite{bhunia2020sketch, song2018learning} dataset.}
    \label{fig:chair1}
\end{figure}

\begin{figure}[!htbp]
    \centering
    \includegraphics[width=0.9\linewidth]{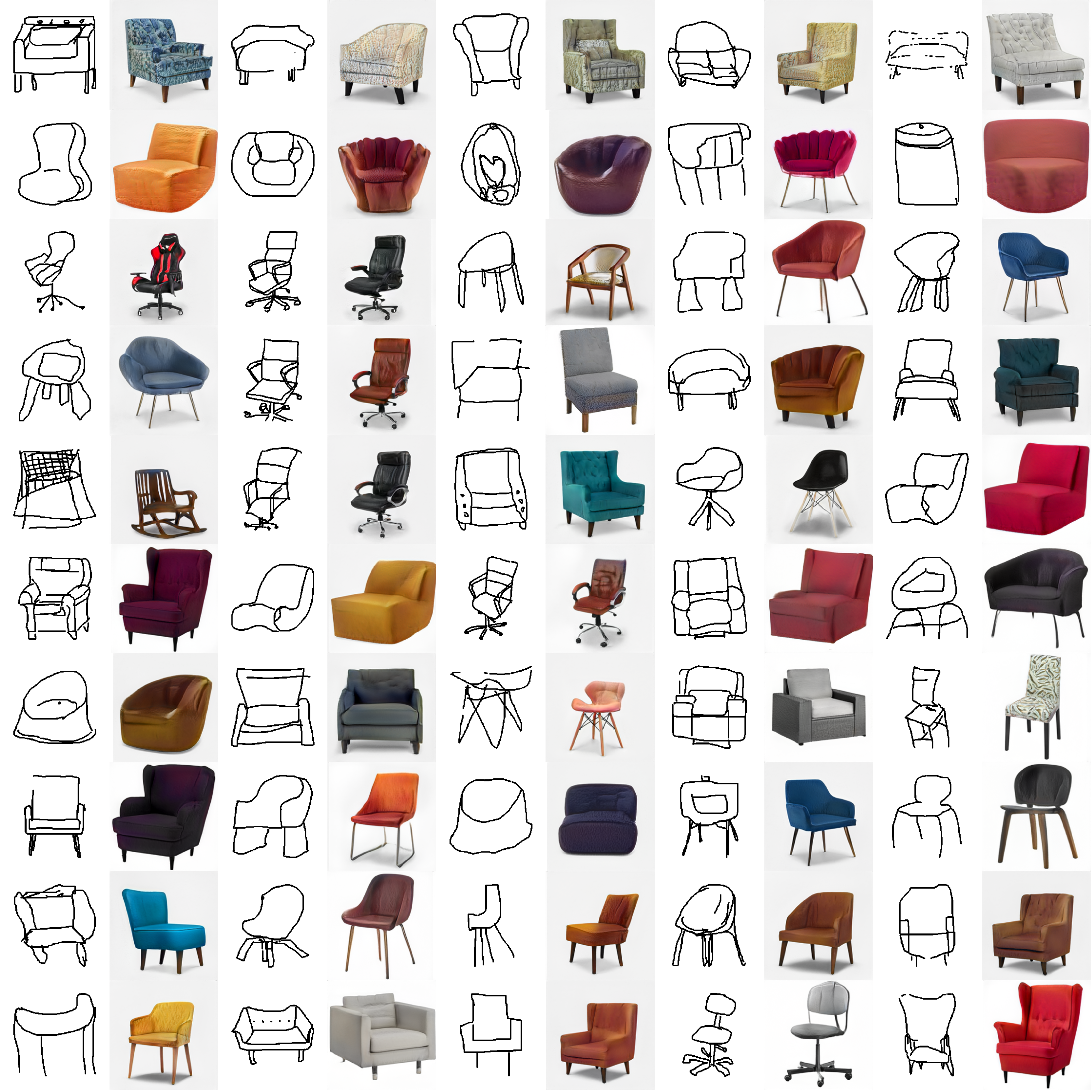}
    \caption{Results generated by the proposed method with sketches from the QMUL-ChairV2~\cite{bhunia2020sketch, song2018learning} dataset.}
    \label{fig:chair2}
\end{figure}

\begin{figure}[!htbp]
    \centering
    \includegraphics[width=0.9\linewidth]{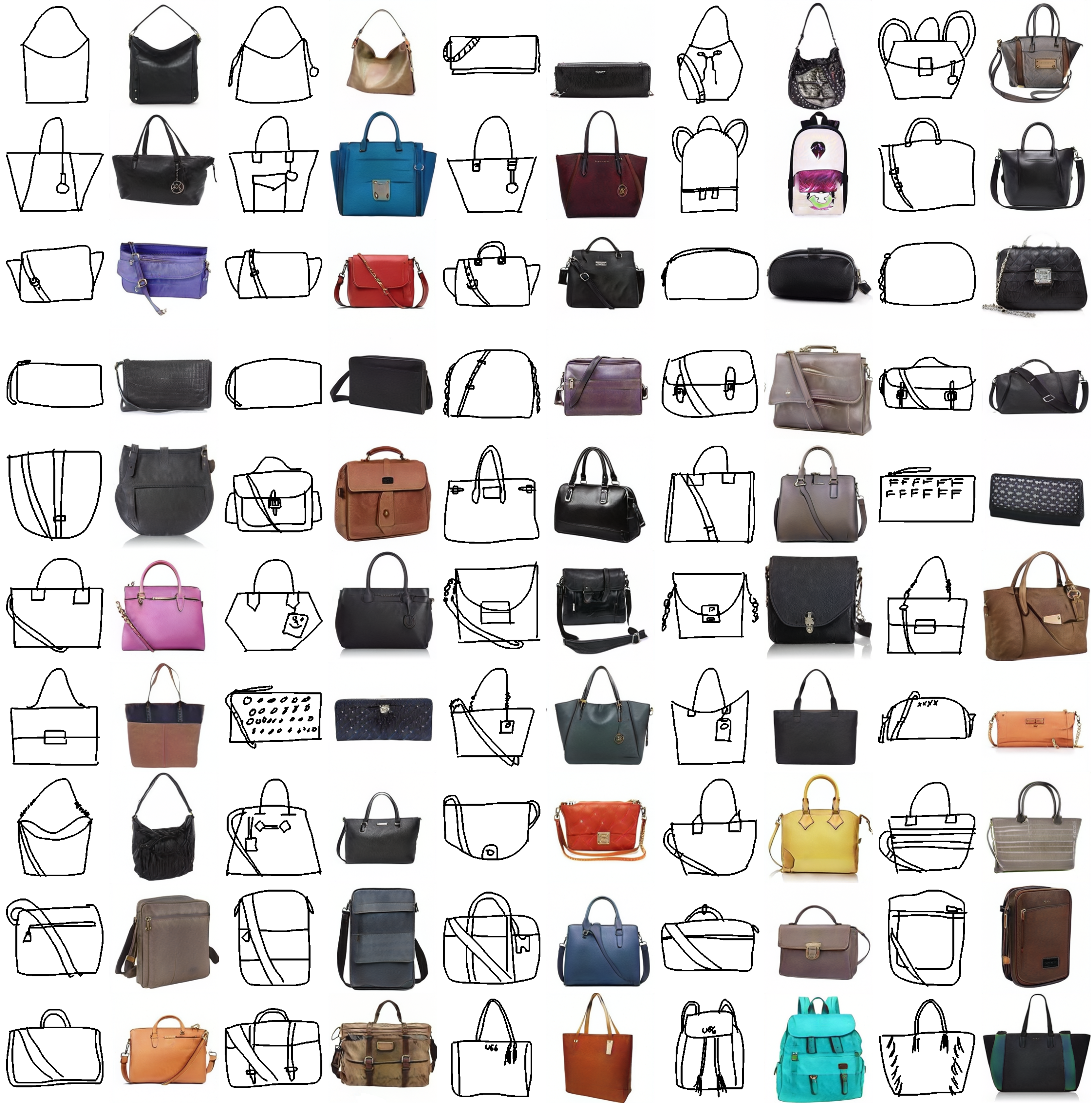}
    \caption{Results generated by the proposed method with sketches from the Handbag~\cite{song2017deep} dataset.}
    \label{fig:handbag1}
\end{figure}

\begin{figure}[!htbp]
    \centering
    \includegraphics[width=0.9\linewidth]{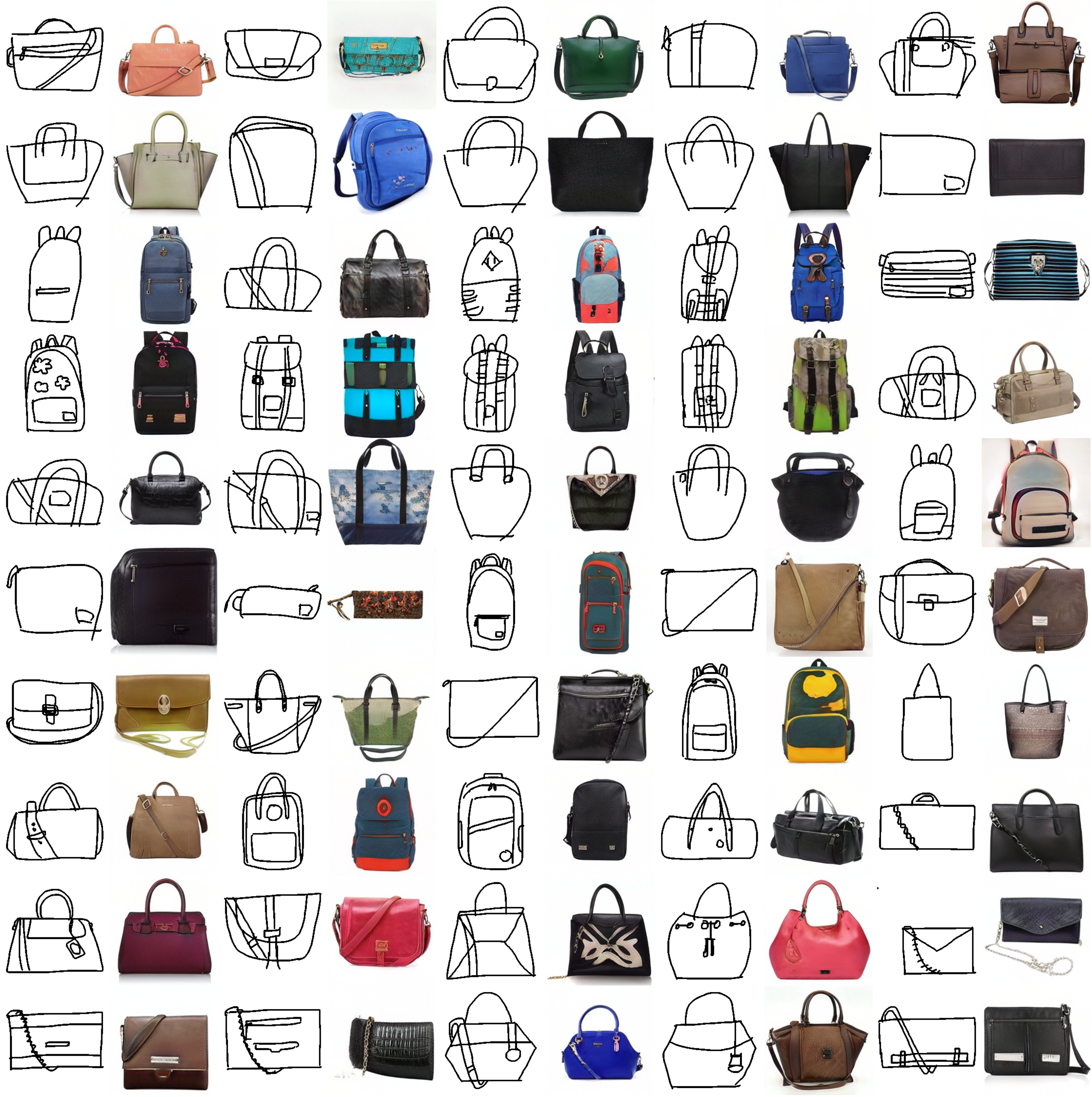}
    \caption{Results generated by the proposed method with sketches from the Handbag~\cite{song2017deep} dataset.}
    \label{fig:handbag2}
\end{figure}

Figs.~\ref{fig:shoe1}-\ref{fig:handbag2} depict additional sketch-to-photo generation results with sketches from QMUL-ShoeV2~\cite{bhunia2020sketch, song2018learning}, QMUL-ChairV2~\cite{bhunia2020sketch, song2018learning}, and Handbag~\cite{song2017deep} datasets. A common observation seen in Figs.~\ref{fig:shoe1}-\ref{fig:handbag2} is that, although photorealistic, the output quality is comparatively lower in ChairV2 and Handbag datasets than that in ShoeV2 due to a higher degree of sketch abstraction and a lower number of training pairs available in the former two datasets.

\clearpage

\section*{B. Details on Human Study}
\begin{figure}[!htbp]
    \centering
    \includegraphics[width=0.77\linewidth]{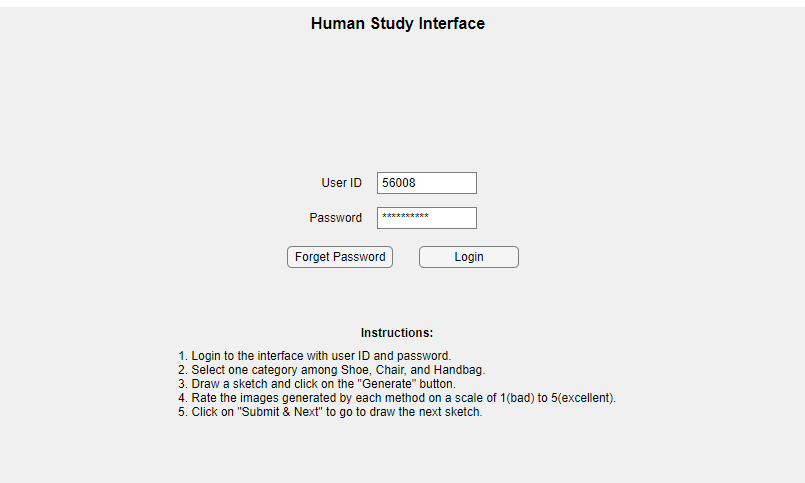}
    \vspace{-0.3cm}
    \caption{Login screen of our human study interface}
    \label{fig:mos_login}
\end{figure}

\begin{figure}[!htbp]
    \centering
    \includegraphics[width=0.77\linewidth]{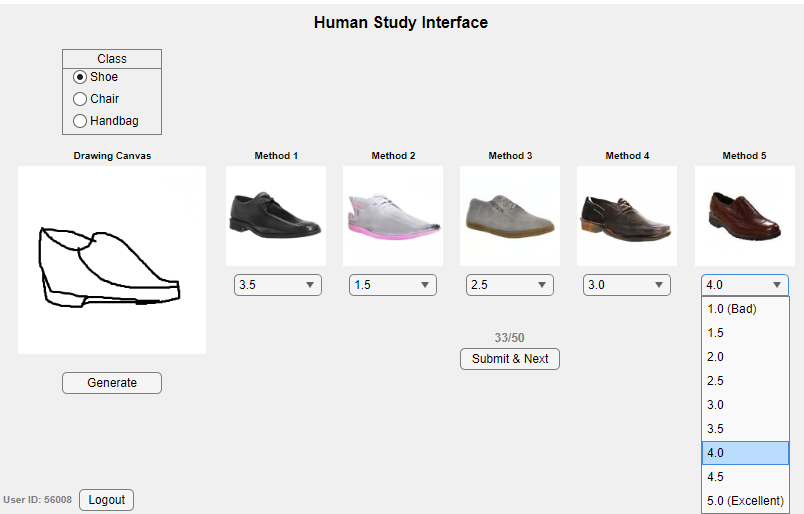}
    \vspace{-0.3cm}
    \caption{Scoring screen of our human study interface}
    \label{fig:mos_scoring}
\end{figure}

\cref{fig:mos_login} and \cref{fig:mos_scoring} depict the login and scoring screen of the interface used to collect the MOS~\cite{huynh2010study} values. Upon login, a user first selects a class (i.e., Shoe, Chair, or Handbag) and draws a sketch of that class. Next, upon clicking on the ``Generate'' button, the system displays corresponding photo translations produced by our proposed method along with every other competing framework. The participant rates every generated photo and clicks on ``Submit \& Next'' to continue. We anonymise the names of the competing methods (e.g., pix2pix~\cite{isola2017image}, MUNIT~\cite{huang2018multimodal}, etc.) to prevent the ratings from being influenced by the participants' past knowledge. For brevity and ease of the participants, following \cite{huynh2010study}, we sub-divide the 1$\rightarrow$5 (bad$\rightarrow$excellent) MOS levels into a nine-point discreet scale with the possible ratings $\{1,1.5,2,2.5,3,3.5,4,4.5,5\}$. For each method, we compute the final MOS~\cite{huynh2010study} value by taking the mean ($\mu$) and variance ($\sigma$) of all individual MOS responses.

\vspace{-0.3cm}

\vspace{-0.6cm}

\clearpage

\section*{C. Intermediate Photo Generation}
\begin{figure}[!htbp]
    \centering
    \includegraphics[width=0.7\linewidth]{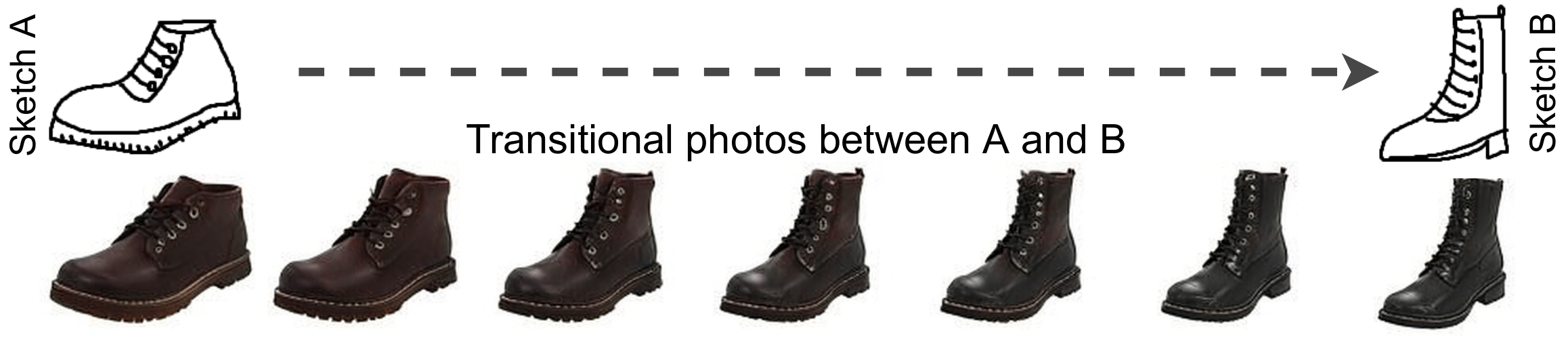}
    \vspace{-0.3cm}
    \caption{Generating transitional shoe photos between Sketch A and B.}
    \label{fig:interpolation1}
\end{figure}

\begin{figure}[!htbp]
    \centering
    \includegraphics[width=0.7\linewidth]{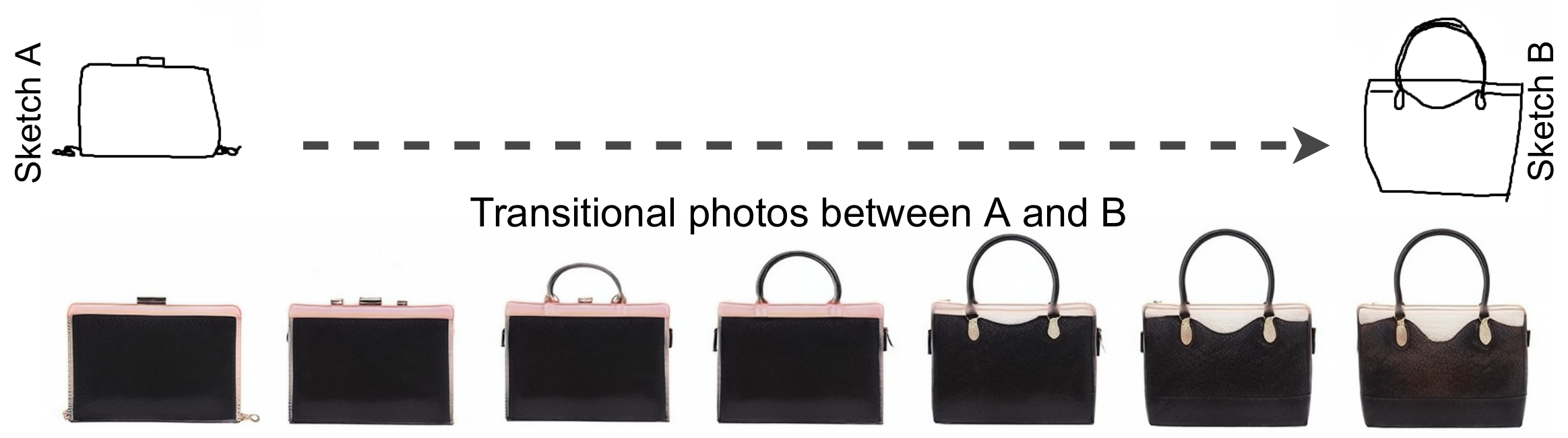}
    \vspace{-0.3cm}
    \caption{Generating transitional handbag photos between Sketch A and B.}
    \label{fig:interpolation2}
\end{figure}

Leveraging StyleGAN's~\cite{karras2020analyzing} smooth latent space, our method allows us to generate realistic transitional photos given an initial and a final sketch. We achieve this via simple arithmetic operations (e.g., interpolation) between the predicted latent codes of the two sketches. \cref{fig:interpolation1} and \ref{fig:interpolation2} shows how given \textit{Sketch A} and \textit{Sketch B}, our method can generate plausible intermediate photos.

\end{document}